\documentclass[12pt]{article}

\usepackage{amsthm}%

\usepackage{newtxtext,newtxmath}

\usepackage{graphicx}

\usepackage[letterpaper, textwidth=18cm, top=18mm, bottom=18mm]{geometry}

\usepackage{comment}
\usepackage{graphicx}%
\usepackage{multirow}%
\usepackage{amsmath}%
\usepackage{newtxmath}%
\usepackage[title]{appendix}%
\usepackage{xcolor}%
\usepackage{textcomp}%
\usepackage{manyfoot}%
\usepackage{booktabs}%
\usepackage{algorithm}%
\usepackage{algorithmicx}%
\usepackage{algpseudocode}%
\usepackage{listings}%

\usepackage[nolist,nohyperlinks]{acronym}
\usepackage{cleveref}
\usepackage{mathtools}
\usepackage{setspace}

\acrodef{DAC}{digital-to-analog converter}
\acrodef{DAgger}{Dataset Aggregation}
\acrodef{DA}{data augmentation}
\acrodef{MPC}{model predictive control}
\acrodef{RTMPC}{robust tube nonlinear model predictive control}
\acrodef{NMPC}{nonlinear model predictive control}
\acrodef{NN}{neural network}
\acrodef{SA}{sampling augmentation}
\acrodef{SQP}{sequential quadratic programming}
\acrodef{UDP}{User Datagram Protocol}
\acrodef{UKF}{unscented Kalman filter}

\newcommand{\Fcmd}{F_{\text{cmd}}}
\newcommand{\taux}{\tau_{x,\text{cmd}}}

\newcommand{\Fdrag}{\mathbf{F}_{\text{drag}}}
\newcommand{\taudrag}{\boldsymbol{\tau}_{\text{drag}}}

\newcommand{\xref}{\mathbf{x}^{\text{ref}}}
\newcommand{\uref}{\mathbf{u}^{\text{ref}}}
\newcommand{\Xref}{\mathbf{X}^{\text{ref}}}
\newcommand{\Uref}{\mathbf{U}^{\text{ref}}}

\newcommand{\Xsetref}{\mathbb{X}^{\text{ref}}}
\newcommand{\Usetref}{\mathbb{U}^{\text{ref}}}
\newcommand{\Xsetmpc}{\mathbb{X}^{\text{mpc}}}
\newcommand{\Usetmpc}{\mathbb{U}^{\text{mpc}}}

\newcommand{\xmpc}{\mathbf{x}^{\text{mpc}}}
\newcommand{\umpc}{\mathbf{u}^{\text{mpc}}}
\newcommand{\Xmpc}{\mathbf{X}^{\text{mpc}}}
\newcommand{\Umpc}{\mathbf{U}^{\text{mpc}}}

\newcommand{\xaug}{\mathbf{x}^{\text{aug}}}
\newcommand{\uaug}{\mathbf{u}^{\text{aug}}}
\newcommand{\Xaug}{\mathbf{X}^{\text{aug}}}
\newcommand{\Uaug}{\mathbf{U}^{\text{aug}}}

\renewenvironment{abstract}
	{\quotation}
	{\endquotation}

\date{}

\makeatletter
\renewcommand{\fnum@figure}{\textbf{Figure \thefigure}}
\renewcommand{\fnum@table}{\textbf{Table \thetable}}
\makeatother

\usepackage{cite}                
\renewcommand\citeleft{\textit{(}}   
\renewcommand\citeright{\textit{)}}  
\renewcommand\citeform[1]{\textit{#1}} 

\usepackage{url}

\title{\textbf{\Large
  Aerobatic maneuvers in insect-scale flapping-wing aerial robots \\
  via deep-learned robust tube model predictive control}}

\author{
  \normalsize
  Yi-Hsuan Hsiao\textsuperscript{1,\dag},
  Andrea Tagliabue\textsuperscript{2,\dag},
  Owen Matteson\textsuperscript{2,\dag},
  Suhan Kim\textsuperscript{1}, \\[4pt]
  \normalsize
  Tong Zhao\textsuperscript{1,2},
  Jonathan P.~How\textsuperscript{2,*},
  YuFeng Chen\textsuperscript{1,*}\\[10pt]
  \footnotesize
  \textsuperscript{1}\,Department of Electrical Engineering and Computer Science, Massachusetts Institute of \\  
  \footnotesize
  Technology, Cambridge, MA 02139, USA\\
  \footnotesize \textsuperscript{2}\,Department of Aeronautics and Astronautics,
  Massachusetts Institute of Technology,\\ 
  \footnotesize Cambridge, MA 02139, USA \\ 
  \footnotesize \textsuperscript{*}Corresponding authors: \texttt{jhow@mit.edu}, \texttt{yufengc@mit.edu}\\
  \footnotesize \textsuperscript{\dag}These authors contributed equally to this work.
}

\begin{document} 

\linespread{0.9}\selectfont
\maketitle
\linespread{1.5}\selectfont
\vspace{-10mm}

\begin{abstract} 
\begingroup
\setstretch{1.2}  
\bfseries \boldmath

Aerial insects exhibit highly agile maneuvers such as sharp braking, saccades, and body flips under disturbance. In contrast, insect-scale aerial robots are limited to tracking non-aggressive trajectories with small body acceleration. This performance gap is contributed by a combination of low robot inertia, fast dynamics, uncertainty in flapping-wing aerodynamics, and high susceptibility to environmental disturbance. Executing highly dynamic maneuvers requires the generation of aggressive flight trajectories that push against the hardware limit and a high-rate feedback controller that accounts for model and environmental uncertainty. Here, through designing a deep-learned robust tube model predictive controller, we showcase insect-like flight agility and robustness in a 750-millgram flapping-wing robot. Our model predictive controller can track aggressive flight trajectories under disturbance. To achieve a high feedback rate in a compute-constrained real-time system, we design imitation learning methods to train a two-layer, fully connected neural network, which resembles insect flight control architecture consisting of central nervous system and motor neurons. Our robot demonstrates insect-like saccade movements with lateral speed and acceleration of 197 centimeters per second and 11.7 meters per second square, representing 447\% and 255$\%$ improvement over prior results. The robot can also perform saccade maneuvers under 160 centimeters per second wind disturbance and large command-to-force mapping errors. Furthermore, it performs 10 consecutive body flips in 11 seconds – the most challenging maneuver among sub-gram flyers. These results represent a milestone in achieving insect-scale flight agility and inspire future investigations on sensing and compute autonomy.

\endgroup

\end{abstract}

\paragraph*{Short Title:} Micro-robotic aerobatics via deep learning.

\paragraph*{One-Sentence Summary:} Biomimetic deep-learned model predictive control enables insect-like aerobatics and flight robustness in a 750-mg robot.

\paragraph*{Main Text:}
\subsection*{INTRODUCTION}\label{sec:intro}

Aerial insects are exquisite flyers capable of performing unique aerobatic feats. Their flight paths consist of straight segments connected by erratic rapid turns, a strategy to localize and navigate in cluttered environments through mitigating vision blur \cite{muijres2015body}. The sharp change of heading direction is called body saccade, and it is characterized by large linear and rotational accelerations up to 10 m/s$^2$ and 3000 rad/s$^2$, respectively \cite{muijres2014flies}. In addition to quickly changing their heading angles, aerial insects can perform fast body flips within 19 ms. This fast evasive maneuver is effective for perching on surfaces and evading predators \cite{muijres2014flies,liu2019flies}.

Inspired by these flight capabilities, flapping-wing aerial robots have been developed to investigate the biomechanics and control principles related to birds and insects. At the mesoscale (20--100 g), several flapping-wing platforms  \cite{tu2020scale,karasek2018tailless} have demonstrated autonomous flight and aggressive maneuvers, such as banked turns \cite{karasek2018tailless} and body flips \cite{tu2021flying}. However, owing to low flapping frequency ($<35$ Hz) and large body inertia ($>2000$ gmm$^2$), these mesoscale robots exhibited substantially slower body dynamics than that of aerial insects. Recent progress in mechanism design and microfabrication led to a new class of sub-gram flapping-wing robots \cite{ma2013controlled,chen2019controlled,chukewad2021robofly,bena2023high,dhingra2025modeling}. These insect-scale systems have demonstrated hovering flight \cite{ma2013controlled}, trajectory tracking \cite{chirarattananon2014adaptive}, multimodal locomotion \cite{chen2017biologically}, damage resilience \cite{mongeau2022flies,kim2023laser}, perching \cite{chirarattananon2016perching}, and body flips \cite{chen2021collision}. Despite showcasing these capabilities, most robotic flights \cite{ma2013controlled,chirarattananon2014adaptive,chen2017biologically,dhingra2025modeling} were characterized by slow flight speeds ($<$40 cm/s) and smooth trajectories lacking rapid turns and high accelerations. In the rare case of performing body flips \cite{kim2025acrobatics}, a state-machine controller decomposed the maneuver into multiple phases where the resultant flight trajectories suffered large positional variance among repeating trials due to unmodeled effects such as tension from the wire tether. There remains a substantial performance gap between aerial insects and their robotic counterparts. 

In our view, the lack of insect-like flight agility is largely contributed by challenges relating to fast dynamics, large model uncertainty, and susceptibility to environmental disturbances. First, sub-gram robots require high-rate feedback controllers owing to small inertia ($<100$ gmm$^2$) and high flapping-wing frequency ($>100$ Hz). Second, it is difficult to accurately predict lift and drag forces associated with flapping-wing propulsion due to unsteady aerodynamic phenomena such as periodic vortex growth and shedding \cite{chen2016experimental}. Very few prior studies investigated the changes of force production under dynamic flight conditions that involve high speed, acceleration, and rotational motion. Third, manufacturing and assembly imprecision often lead to large model uncertainty that requires painstaking system identification and parameter tuning. Fourth, low-inertia insect-scale robots are easily influenced by environmental disturbance, such as ambient airflow and tension forces from the power tether. Thus, executing insect-like fast saccades and body flips requires a computationally efficient flight controller that can plan complex trajectories and handle model uncertainty. 

\begin{figure}[h]
\centering
\includegraphics[width=180mm]{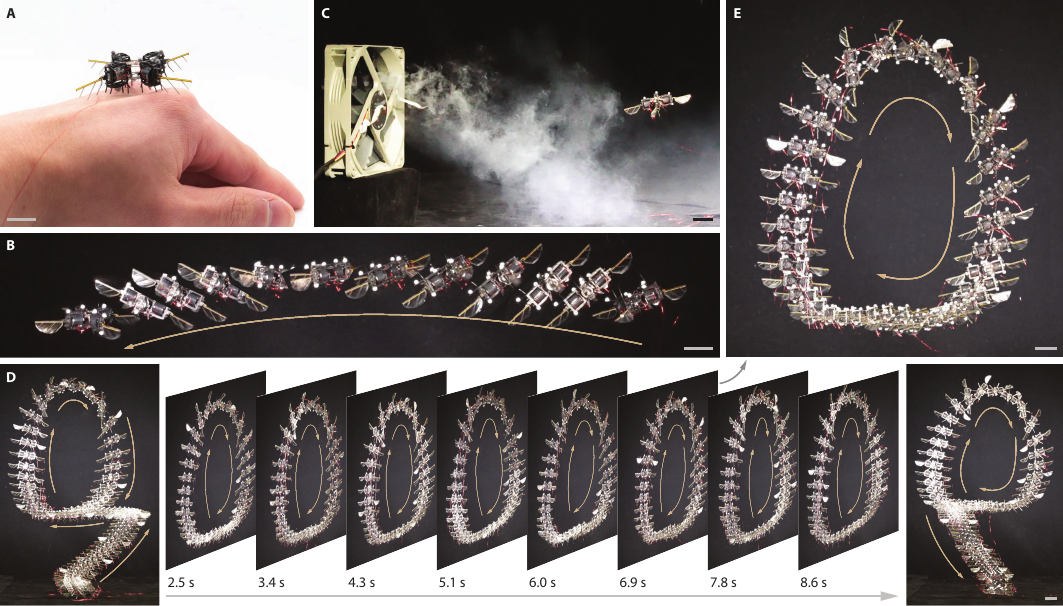}
\vspace{-8mm}
\caption{\textbf{Overview of flight maneuvers performed by a 750 mg flapping-wing aerial robot.} (\textbf{A}) An image of the robot resting on the back of hand. (\textbf{B}) A composite image of an insect-like body saccade maneuver. (\textbf{C}) The robot tracked a repeated saccade trajectory under 160 cm/s wind disturbance. (\textbf{D}) A time sequence of composite images that illustrate ten consecutive body flips in an 11-s flight. (\textbf{E}) An inset image of (D) that illustrates the seventh flip in the flight experiment. Scale bars in (A-E) represent 1 cm.}\label{fig:overview}
\end{figure}

In this work, we constructed a deep-learned control policy from a robust tube nonlinear model predictive controller (RTMPC) \cite{mayne2011tube} to execute insect-like agile maneuvers in a 750-mg flapping-wing robot (Fig. \ref{fig:overview}A and movie S1). An expert \acs{RTMPC} could track aggressive trajectories under large model uncertainty and environmental disturbance, and it was used to train a neural network (NN) controller via data-efficient imitation learning (IL). The advantage of this approach was that it combined high performance and computational efficiency, leading to high-rate execution of numerous challenging flight maneuvers. Specifically, our robot imitated insect-like body saccades through following straight paths interspersed with rapid turning, which featured large acceleration and aggressive body pitching (Fig. \ref{fig:overview}B). The robot’s maximum flight speed and acceleration were 197 cm/s and 11.7 m/s$^2$ — representing 447\% and 255\% increase over the state-of-the-art trajectory tracking result \cite{kim2025acrobatics}. In addition to maintaining best-in-class tracking accuracy when performing these never-attempted trajectories, our robot further showcased robustness against 160 cm/s wind gust (Fig. \ref{fig:overview}C) and 33\% error in command-to-thrust mapping. Furthermore, it demonstrated ten consecutive body flips (Fig. \ref{fig:overview}D-E) in an 11-second flight, which highlighted unprecedented agility, precision, and robustness in aerial robotic insects. Although our controller was run on an offboard computer, we analyzed the tradeoff between computation requirement and flight performance. Through reducing the NN size in the IL procedure, we envision our controller design will enable compute-autonomous flight in future sub-gram aerial systems.

\subsection*{RESULTS}\label{sec:results}

\subsubsection*{Robot and flight controller design}
We built an insect-scale flapping-wing robot (Fig. \ref{fig:overview}A) based on a recent design \cite{kim2025acrobatics}. The robot had a dimension of 4 cm $\times$ 4 cm $\times$ 0.9 cm, weighed 750 mg, and consisted of four independent flapping-wing modules each driven by a dielectric elastomer actuator (DEA). In this work, the robot relied on offboard motion capture systems, control computers, and power sources (fig. S1 and Supplementary Text). The robot was equipped with seven 1.5-mm reflective markers that were tracked by six motion capture cameras (Vantage V5, Vicon) at approximately 400 Hz. The position and attitude data were sent to a computer (Speedgoat Baseline) that controlled four high voltage amplifiers. The amplifiers output sinusoidal signals with a fixed frequency (330 Hz) and time-varying voltage amplitudes. Consequently, the robot flapped its wings at a constant frequency of 330 Hz while it modulated its flapping-wing amplitudes. 

This hardware design posed unique and critical control challenges because the robot had low inertia and high flapping frequency. Due to its small size and mass, the robot was susceptible to environmental disturbances such as ambient flow and tension from its power tether. The robot was also intrinsically unstable, requiring high-rate feedback control to stabilize the attitude. Most prior work \cite{ma2013controlled,chen2019controlled,kim2025acrobatics} adopted model-based controllers that were computationally simple, which have only demonstrated low speed ($<$40 cm/s) flights far inferior to those of aerial insects. To the best of our knowledge, existing flight controllers for insect-scale robots were unable to simultaneously address the requirement on high feedback rate and the goal of achieving agile flights under large uncertainty and disturbances.    

To address the dilemma between performance and computational cost, we proposed a two-stage controller design. Our flight controller employed a computationally efficient NN policy that was trained on a high-performing RTMPC, which was being treated as a demonstrator. In the first stage, the expert \acs{RTMPC} (Fig. \ref{fig:controller}A) was designed to optimize robot flight performance without considering computational cost. It consisted of a safe reference trajectory generator, an ancillary nonlinear MPC, and a sampling method for constructing the convergence tube (Fig. \ref{fig:controller}B). First, the trajectory generator computed a dynamically feasible reference path $\Xref$ and its corresponding command $\Uref$ with tightened constraints. Second, the ancillary nonlinear MPC tracked the planned trajectory over a finite receding horizon and recorded the state sequence $\Xmpc$ and command sequence $\Umpc$. Compared to classical controllers \cite{kim2025acrobatics}, the nonlinear MPC considered force and torque constraints – critical for tracking aggressive trajectories that required commands close to the hardware limit. In contrast to prior implementations \cite{tagliabue2023robust,tagliabue2024efficient} that employed a cascaded architecture consisting of a geometric attitude controller and an MPC position controller, our design involved a single loop MPC that combined attitude and position control. This design directly output force and torque commands under hardware constraints at the expense of higher computational cost. Lastly, a disturbance-invariant tube was constructed via Monte-Carlo simulations of the nominal system subject to sampled uncertainty realizations. This tube construction ensured the robot could follow the desired trajectory if its states resided within the bounded region by a distance $\Delta \mathbf{x}$ (See Materials and Methods: Robot Dynamic Model, Safe reference trajectory generator, Nonlinear ancillary model predictive control, and Tube construction for implementation details). 

In the second stage, we trained a computationally efficient NN policy $\pi$ based on the expert \acs{RTMPC} via IL. The training process required many state and action pairs sampled around the planned trajectory, and it would be prohibitively expensive to solve a \ac{SQP} for each sample. To train the NN policy efficiently, we employed a data augmentation method named \ac{SA} \cite{tagliabue2024efficient}, which leveraged the safety tube constructed from stage 1 and sampled additional state–action pairs within this tube. This method utilized the sensitivity matrix $\mathbf{K}$ computed by the ancillary nonlinear MPC at each timestep. Based on $\mathbf{K}$ and the perturbed state $\Xaug$ within the tube, we efficiently calculated the augmented command $\Uaug$ that was close to the nonlinear MPC action $\Umpc$. Next, we trained an NN control policy based on this sample-efficient method. The initial policy relied on both the nonlinear ancillary MPC's $\Xmpc$ and $\Umpc$ and the augmented state-action pairs $\Xaug$ and $\Uaug$. After obtaining the initial policy, we took very few demonstrations from the RTMPC teacher policy and performed policy fine-tuning through \ac{DAgger} \cite{ross2011reduction}. The implementation details of the sampling and learning methods are described in Materials and Methods: Data augmentation via parametric sensitivities and Imitation learning of neural network controller.  



The resultant NN took robot states and time as input, was composed of two fully connected neuron layers, and returned the control commands (Fig. \ref{fig:controller}C). The inputs were prescribed by robot position $\mathbf{p}$, velocity $\mathbf{v}$, attitude $\mathbf{q}$, angular velocity $\boldsymbol{\omega}$, and time $t$. The output commands were thrust force $\Fcmd$ and torques $\taux$ and $\taux$. The output force and torques were later mapped into driving voltage for each robot module (Supplementary Text). This deep-learned RTMPC policy was central to our controller implementation and each loop update only required a few microseconds on a desktop CPU (Intel-i3-5005U). This NN policy was particularly suitable for insect-scale aerial robots with ultrafast dynamics because it maintained RTMPC-level performance under a substantially lower computation cost. Fig. \ref{fig:controller}C illustrates the high-level control system architecture that consisted of a motion capture setup for measuring the robot position and attitude, an unscented Kalman filter (UKF) for estimating full robot states and environmental disturbance, the deep-learned RTMPC for generating commands, an allocation matrix for mapping the force and torque commands into driving signals for each robot module, and the robot hardware. The UKF implementation is described in Materials and Methods: Unscented Kalman Filter.

\subsubsection*{Body saccade demonstrations under uncertainty and disturbance}
\begin{figure}[h]
\centering
\includegraphics[width=180mm]{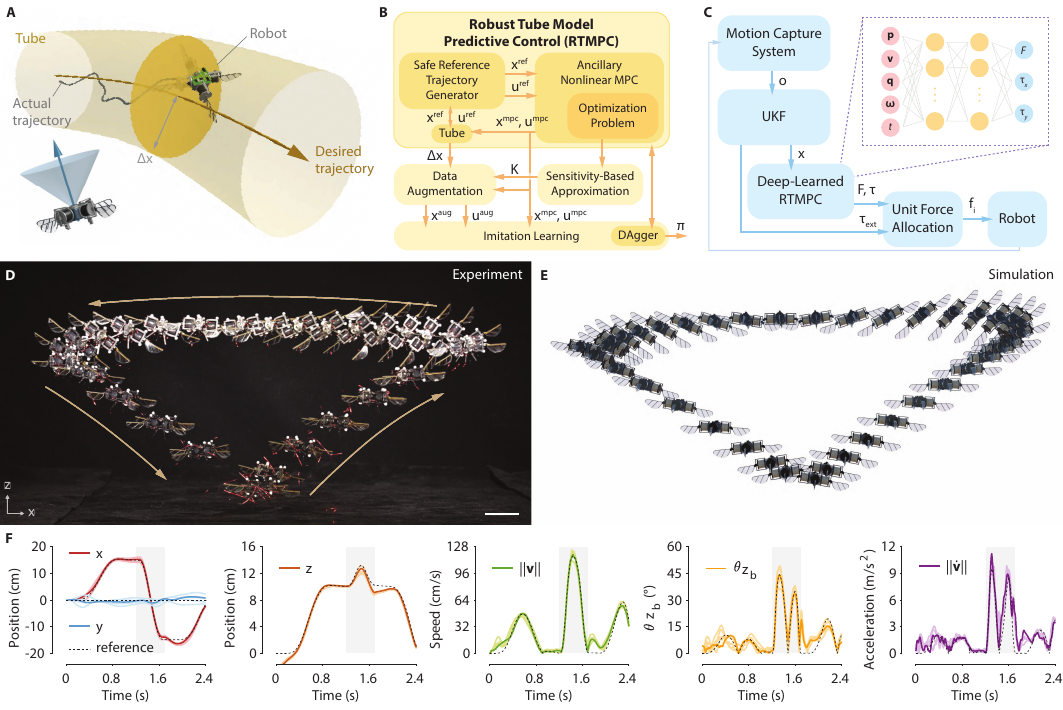}
\vspace{-8mm}
\caption{\textbf{Controller design and body saccade demonstration.} (\textbf{A}) Illustration of a robust tube model predictive controller (RTMPC) for tracking aggressive flight trajectories under disturbance and uncertainty. (\textbf{B}) A deep-learned policy is trained based on the RTMPC to enable real-time deployment on robot hardware. (\textbf{C}) The closed-loop control architecture consists of a motion capture system, an unscented Kalman filter (UKF), a deep-learned RTMPC controller, robot force and torque mapping, and robot hardware. (\textbf{D}) A composite image of robot takeoff, body saccade, and landing. Scale bar represents 1 cm. (\textbf{E}) A simulated saccade maneuver closely resembled the flight experiment in (D). (\textbf{F}) Tracked robot position, speed, body deviation angle, and acceleration corresponding to the flight in (D). The lighter-colored curves show data from repeated experiments. The shaded region highlights the dynamic maneuver during which performance metrics (speed, positional error) are computed from.}\label{fig:controller}
\end{figure}

We quantified the robot and controller performance by executing insect-like body saccade movements. In contrast to following smooth trajectories, aerial insects make frequent sharp braking and turning between random setpoints. This erratic flight strategy allows them to confuse and evade predators, as well as reducing visual motion blur when navigating in cluttered environments. This flight pattern is difficult to replicate because it involves large lateral acceleration and deceleration, which further requires large pitching and rolling motion exceeding 45° \cite{muijres2014flies}. Fig. \ref{fig:controller}D and movie S2 illustrate fast movement between two setpoints. The robot ascended upward to the right-hand-side setpoint, then quickly accelerated laterally before aggressively braking at the left-hand-side setpoint, and finally descended to land at the takeoff location (Fig. \ref{fig:controller}D). This flight demonstration closely resembled the simulated maneuver (Fig. \ref{fig:controller}E and Supplementary Text), which highlighted exceptional controller performance. During the lateral flight segment (shaded region in Fig. \ref{fig:controller}F), the robot moved across 30 cm within 0.47 s and reached a maximum speed of 124 cm/s. In the acceleration and deceleration phases, the robot pitched aggressively (Fig. \ref{fig:controller}D) where the maximum body deviation angle and acceleration reached 49.1° and 11.4 m/s$^2$, respectively (Fig. \ref{fig:controller}F). In this maneuver, the robot acceleration, body deviation angle, and speed exceeded prior results \cite{ma2013controlled,kim2025acrobatics} by 245\%, 182\%, and 243\%, respectively. We repeated this maneuver five times (fig. S2) and measured root mean square (RMS) lateral and altitude error of 1.12 cm and 0.75 cm compared to the desired trajectory. 

\begin{figure}[h]
\centering
\includegraphics[width=180mm]{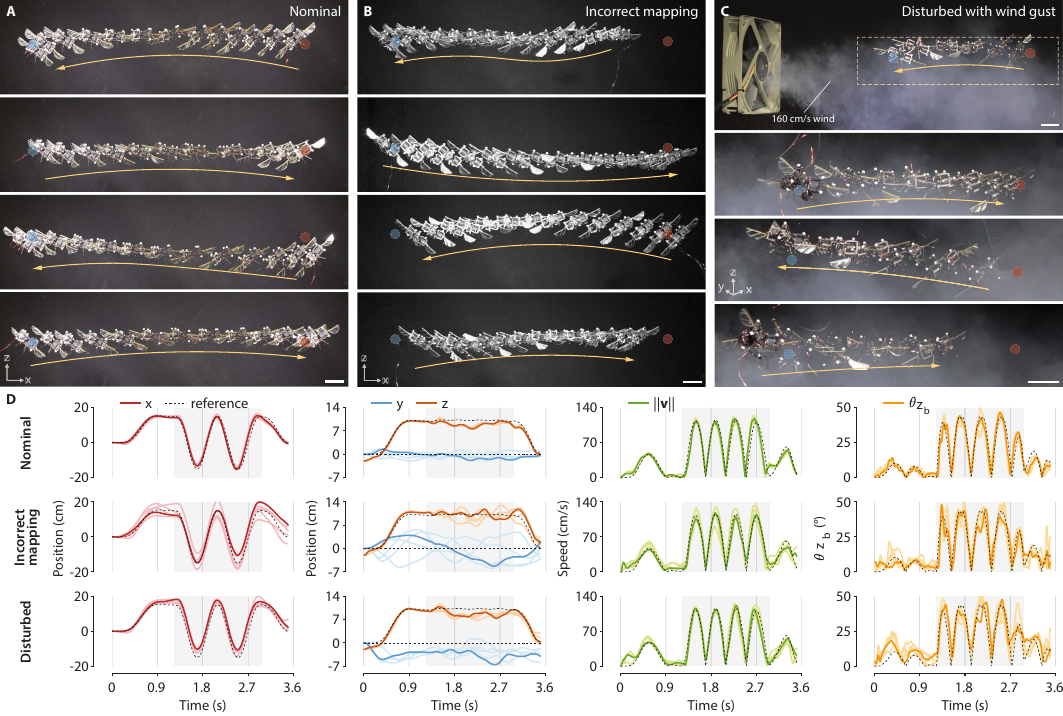}
\vspace{-8mm}
\caption{\textbf{Repeated saccade demonstrations under mapping error and wind disturbance}. (\textbf{A}) A composite image of repeated body saccade between two setpoints. (\textbf{B}) A composite image of the same experiment conducted under 33\% command-to-force mapping error. (\textbf{C}) A composite image of the same experiment under 160 cm/s wind disturbance. The camera is placed at a 45° angle relative to the flight plane.
(\textbf{D}) Tracked robot position, speed, and body deviation angle. The three rows correspond to the flights in (A-C), respectively. The lighter-colored curves show data from repeated experiments. Scale bars in (A-C) represent 1 cm.} 
\label{fig:multi_p2p}
\end{figure}

Based on this result, we performed repeated body saccade maneuvers where the robot switched between a pair of setpoints four times (Fig. \ref{fig:multi_p2p}A-C and movie S3). This trajectory evaluated controller reliability under repeated aggressive commands and the robot’s ability to make 180° sharp turns. Fig. \ref{fig:multi_p2p}D shows the tracked flight data of five flight experiments (fig. S3) in an undisturbed setting (Fig. \ref{fig:multi_p2p}A) and compares them with the desired trajectory. The robot lateral and altitude RMS errors were 2.58 cm and 1.28 cm while it achieved a maximum speed and body deviation angle of 124 cm/s and 50.0°, respectively. This result showed similar performance compared to the simpler maneuver in Fig. \ref{fig:controller}D. 

We further evaluated controller robustness against model uncertainty and environmental disturbance. First, we conducted the same flight experiments while injecting a 33\% error in the command-to-force mapping (Fig. \ref{fig:multi_p2p}B, movie S3, and fig. S4). Despite experiencing different mapping errors in each of the four flapping-wing modules, the robot completed the trajectory with 4.72 cm RMS positional error (second row in Fig. \ref{fig:multi_p2p}D) -- almost 2 times that of the well-calibrated case (first row in Fig. \ref{fig:multi_p2p}D). This result showed that the robot could track aggressive trajectories with moderate accuracy under a large calibration error. It further opened opportunities for operating multiple robots without requiring meticulous system identification.

In addition, we conducted the same flights under 160 cm/s wind disturbance, which was comparable to the robot's maximum sustained fight speed (Fig. \ref{fig:multi_p2p}C-D, movie S3, and fig. S5). The incident air flow and the high-speed camera were obliquely aligned at 45° with respect to the lateral flight segment (fig. S1D), and we injected fine glycerin particles to visualize the wind gust. Without prior knowledge of wind disturbance, the controller increased commanded forces and torques to reject disturbance, resulting in 60\% higher mean body angular velocity (fig. S5). The robot completed five flights with 4.58 cm RMS positional error – a mere 58.5\% increase compared to the flights done in still environments (Fig. \ref{fig:multi_p2p}D). A prior result on hovering flight disturbance rejection reported 2.4 cm RMS error under 60 cm/s wind speed \cite{chirarattananon2017dynamics}. Our result achieved comparable position accuracy while the robot performed aggressive maneuvers under 267\% of the wind disturbance. These flight experiments highlighted controller robustness under large uncertainty and disturbance, which was crucial to perform more challenging trajectories that involved unmodeled aerodynamic and dynamic effects.          

\subsubsection*{Aggressive flight maneuvers characterized by large acceleration and speed}
\begin{figure}[h]
\centering
\includegraphics[width=180mm]{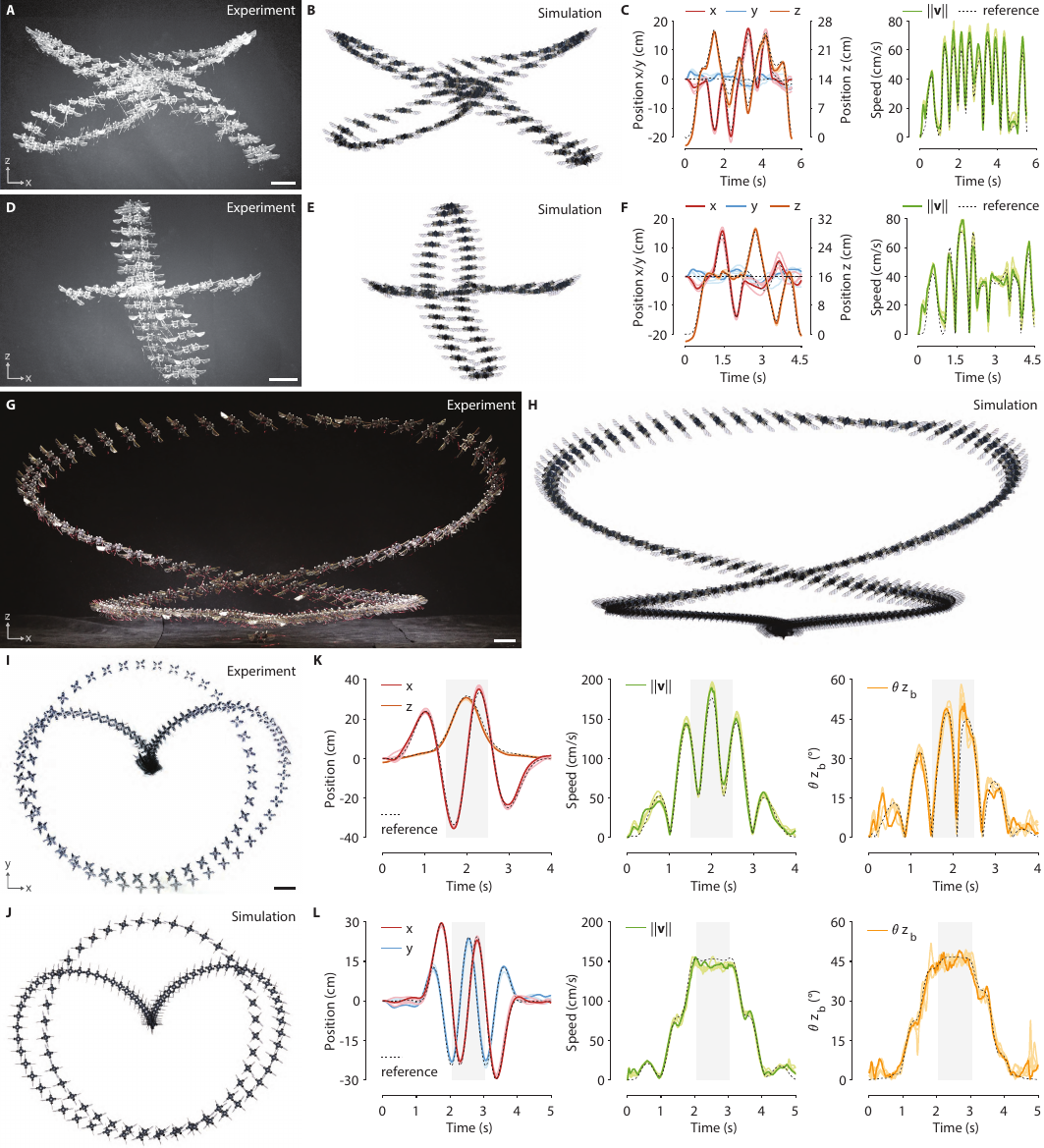}
\vspace{-4mm}
\caption{\textbf{Tracking of aggressive flight trajectories.} (\textbf{A}) A composite image of an X-shaped trajectory that involves nine sharp turns. (\textbf{B}) A composite image of the same maneuver (A) in simulation. (\textbf{C}) Tracked robot position and flight speed corresponding to the flight in (A). (\textbf{D}) A composite image of a cross pattern that involves eight sharp turns and fast vertical descent under unmodeled downwash. (\textbf{E}) A composite image of the same maneuver (D) in simulation. (\textbf{F}) Tracked robot position and flight speed corresponding to the flight in (D). (\textbf{G}) A side-view composite image of the robot tracking a figure 8 trajectory. (\textbf{H}) A composite image of the same maneuver (G) in simulation. (\textbf{I}) A top-view composite image of the robot tracking a planar circular trajectory. (\textbf{J}) A composite image of the same maneuver (I) in simulation. (\textbf{K}-\textbf{L}) Tracked robot position, speed, and body deviation angle corresponding to the flights shown in (G) and (I), respectively. In (K) and (L), the shaded regions indicate high speed flight segments where the performance metrics are computed. The lighter-colored curves in (C, F, K, and L) show data from repeated experiments. Scale bars in (A, D, G, and I) represent 3 cm. }\label{fig:long_traj}
\end{figure}

The flight experiments illustrated in Fig. \ref{fig:controller} and Fig. \ref{fig:multi_p2p} consisted of horizontal linear segments. To evaluate robot agility and precision in tracking complex trajectories, we designed four flight paths characterized by frequent braking, sharp turning, and high flight speed. The first trajectory (Fig. \ref{fig:long_traj}A-C and movie S4) resembled an X-shaped path on the xz-plane. The robot made nine abrupt turns in 5.5 seconds with a mean flight speed of 41.1 cm/s. This demonstration was repeated five times (fig. S6) with RMS lateral and altitude error of  2.11 cm and 1.24 cm, respectively. The second trajectory (Fig. \ref{fig:long_traj}D-F and movie S4) was composed of horizontal saccades and fast vertical movements. Comparing the five repeated experiments (Fig. \ref{fig:long_traj}F and fig. S7), we observed larger positional variance in the fast descent phase because unmodeled downwash adversely reduced robot thrust. The controller compensated for this unmodeled effect by commanding 4.72\% higher lift force than that in simulation (Fig. \ref{fig:long_traj}E). The robot performed this trajectory with an average speed of 40 cm/s and RMS lateral and altitude error of 2.72 cm and 0.74 cm, respectively. 


\newpage
\newgeometry{top=8mm, bottom=16mm, left=15mm, right=15mm}

\newpage
\restoregeometry

While the first two trajectories featured sharp braking and turning, each flight segment was short ($<$30 cm) and the robot could not accelerate to a high speed. We designed and evaluated two trajectories (Fig. \ref{fig:long_traj}G-J) with longer smooth segments for the robot to accelerate for longer durations. Fig. \ref{fig:long_traj}G-H demonstrated a trajectory resembling the figure-8 shape along the xz-plane (Fig. \ref{fig:long_traj}G-H and movie S5). The upper half of the flight pattern spanned 70 cm by 20 cm, allowing the robot to accelerate over a larger distance. During this flight segment (shaded region in Fig. \ref{fig:long_traj}K), the robot’s maximum and average speeds were 197 and 109 cm/s, with the maximum representing 446\% improvement over the fastest result among sub-gram robots \cite{kim2025acrobatics}. The robot repeated this flight five times (fig. S8) with RMS lateral and altitude error of 2.60 cm and 1.33 cm, respectively. In another demonstration, the robot followed a circular trajectory along the xy-plane (Fig. \ref{fig:long_traj}I and movie S5). The robot took off from the center of the circle and then accelerated toward the circle perimeter via a curved path. Next, it followed the circle perimeter at 152 cm/s (shaded region in Fig. \ref{fig:long_traj}L) before decelerating and landing at the takeoff position. The robot repeated this maneuver five times (fig. S9) with RMS lateral and altitude error of 1.80 cm and 1.84 cm, respectively. Both demonstrations were characterized by large body deviation angles and high flight speeds (Fig. \ref{fig:long_traj}K and L), which highlighted substantial improvement of flight agility. These demonstrations (Fig. \ref{fig:long_traj}G and I) also closely resembled the simulated trajectories (Fig. \ref{fig:long_traj}H and J), featuring exceptional tracking accuracy in highly dynamic flights. Based on our simulations, we estimated the robot’s maximum flight speed could exceed 300 cm/s but it was limited by the size of motion capture arena and the unmodeled tension and weight of the robot’s power tether.

\subsubsection*{Aerobatic body flips}
\begin{figure}[H]
\centering
\includegraphics[width=180mm]{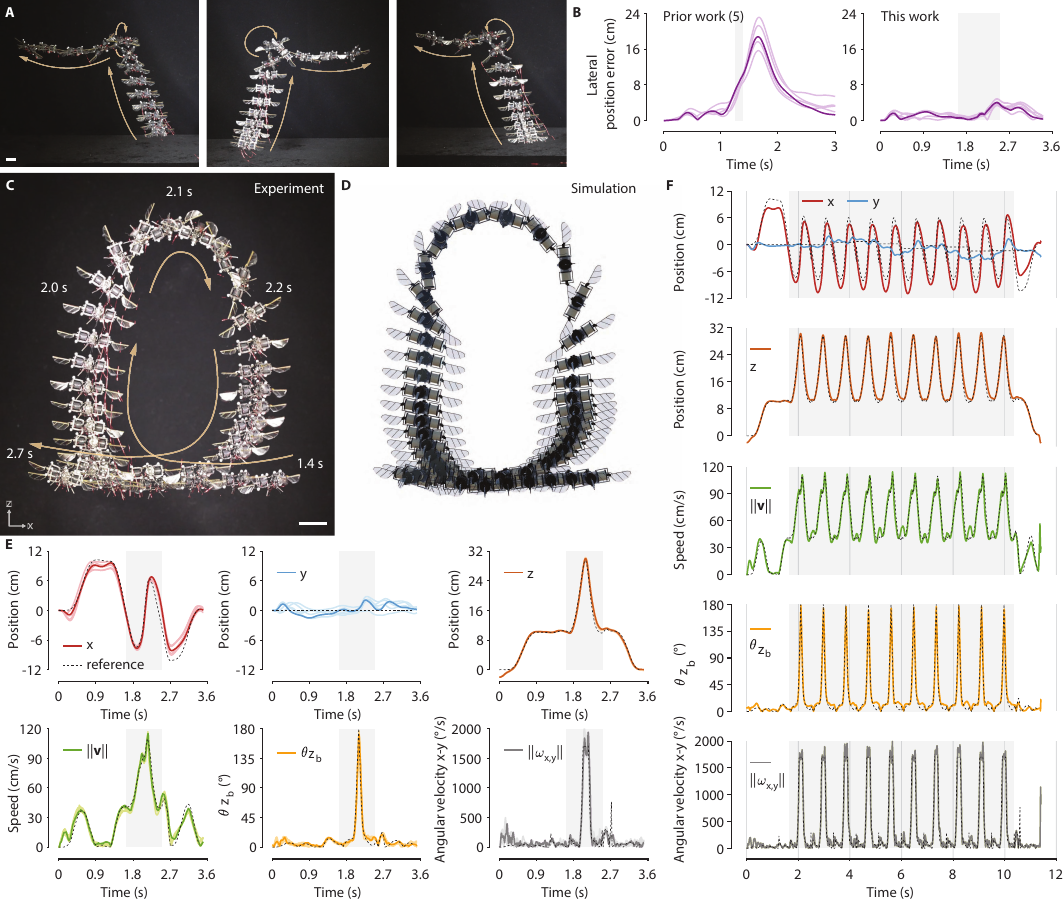}
\vspace{-8mm}
\caption{\textbf{Body flip demonstrations}. (\textbf{A}) Composite images of body flip experiments performed with a prior controller. (\textbf{B}) Body flip position error comparison between the prior and this work. (\textbf{C}) A composite image of one complete flip performed under the deep-learned RTMPC controller. (\textbf{D}) A composite image of the same maneuver (C) in simulation. (\textbf{E}) Tracked robot position, speed, body deviation angle, and angular speed corresponding to (C). (\textbf{F}) Tracked flight data of the ten body flips experiment that correspond to the demonstration shown in Fig. \ref{fig:overview}D. The shaded regions in (B), (E), and (F) indicate durations where the robot performs body flips.  The lighter-colored curves in (B) and (E) show data from repeated experiments. Scale bars in (A) and (C) represent 1 cm.}\label{fig:flip}
\end{figure}

Aerobatic body flips represented the most challenging maneuver for sub-gram aerial robots because they required rapid changes of thrust forces and large translational and angular accelerations under unmodeled disturbances. Of the very few previous works \cite{chen2021collision,kim2025acrobatics} that attempted flips in sub-gram aerial robots, the maneuvers were manually decomposed into multiple stages involving ascent, partially open-loop flip, and attitude recovery. This approach suffered large positional variance (Fig. \ref{fig:flip}A-B) and a low success rate. Fig. \ref{fig:flip}A shows a composite image of three body flip experiments conducted under the old controller \cite{kim2025acrobatics}, and Fig. \ref{fig:flip}B highlights the large positional error of each trial. This lack of maneuver accuracy limited the robot’s ability to track complex trajectories composed of a sequence of aggressive movements. 

Our control approach substantially improved positional accuracy and reliability. Fig. \ref{fig:flip}C and movie S6 show a body flip maneuver that closely resembled the simulated trajectory (Fig. 5D). This experiment was repeated five times (fig. S10), and the flights showed 2.24 cm RMS positional error relative to the desired flight path. As indicated by the shaded regions in Fig. \ref{fig:flip}E, the robot’s maximum flight speed and angular velocity reached 115 cm/s and 2030 °/s, highlighting remarkable flight precision when it performed some of the most challenging maneuvers. This high positional accuracy allowed repetition of challenging maneuvers in the same flight. The robot performed flights where it conducted three, five, seven, and ten consecutive body flips (Fig. \ref{fig:flip}F, fig. S11 and movie S6). Our robot maintained low RMS positional error of 2.49 cm in the ten-flip experiment (Fig. \ref{fig:overview}D, Fig. \ref{fig:flip}F, and movie S6), which highlighted exceptional controller robustness. This demonstration was particularly challenging due to the highly nonlinear dynamics and strong coupling between robot's translation and rotation. During the flip, position control was severely underactuated, and the maneuver operated near actuator limits. It also required precise torque application and fast control execution on the tens of millisecond scale. In addition, the power tether occasionally wrapped over the robot during the flipping process. As shown in movie S6, the tether entangled around the robot airframe after the second and sixth flips. In these two flipping cycles, the robot maintained good flight precision despite experiencing repeated wing-tether collisions. This highlights the strengths of the proposed deep-learned RTMPC framework, which accounts for nonlinear dynamics, external disturbances, and actuator constraints during controller synthesis. As a result, the NN policy executed quickly and remained robust even under unmodeled interactions, such as tether dynamics encountered during the flips.

Our tube controller design also ensured flight safety. During flight, the power tether could become entangled and abruptly exert large forces and torques on the robot. While it was difficult to guarantee NN's response if the states went beyond the trained dataset, our controller could check whether the robot states remained within the tube (Supplementary Text). If the robot states deviated outside the tube, our system would immediately switch on a model-based controller that stabilizes and lands the robot. An example of anomalous state detection and controller switching was demonstrated in fig. S12 and movie S7. This design ensured our robot could reliably perform aggressive maneuvers under large disturbance without risking hard crashes.   

\newpage

\subsection*{DISCUSSION}

\begin{figure}[h]
\centering
 \includegraphics[width=180mm]{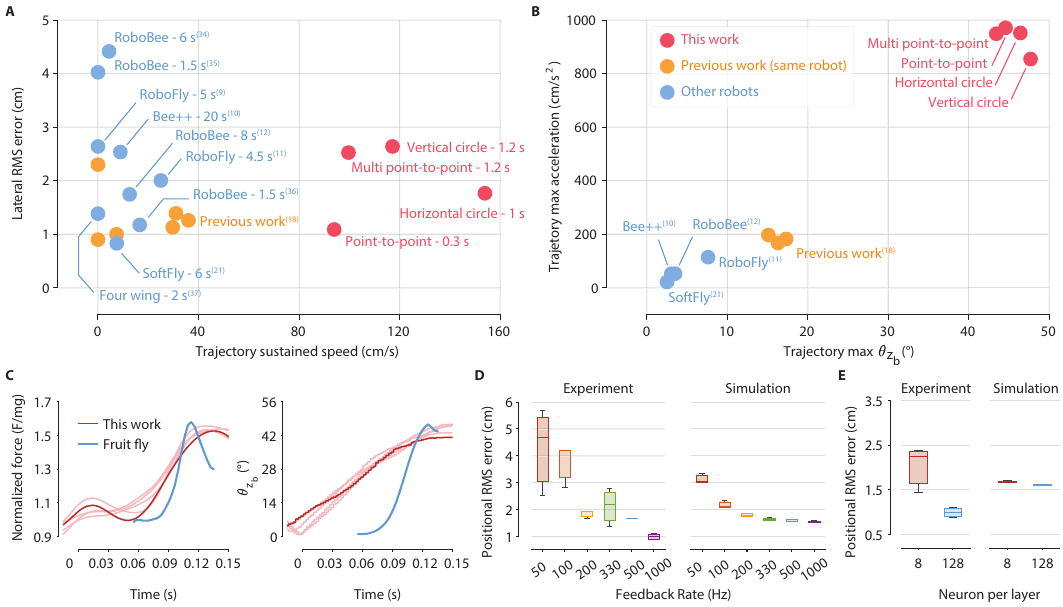}
\vspace{-8mm}
\caption{\textbf{Comparison of robot flight performance}. (\textbf{A-B}) Robot flight precision, speed, instantaneous acceleration, and body angle deviation comparison with that of existing sub-gram aerial robots. (\textbf{C}) Comparison of robot and fruit fly thrust force production and body tilting angle during saccade maneuvers. The blue lines show insect flight data from a prior study \cite{muijres2014flies}. (\textbf{D}) In both experiment and simulation, flight accuracy was inversely correlated with controller feedback rate. (\textbf{E}) At the expense of higher positional error, the robot could perform body saccade movement with substantially smaller number of neurons in the NN. The data corresponding to each flight experiment in (D-E) is shown in fig. S13-S16.}\label{fig:comparison}
\end{figure}

In this work, we developed a deep-learned RTMPC flight controller for demonstrating insect-like agile and robust maneuvers in a sub-gram flapping-wing robot. Our robot could perform aggressive body saccades, track complex trajectories, resist large wind disturbances, and repeatedly execute body flips with high positional accuracy. Compared to existing sub-gram robots, our work achieved over 446\% increase in flight speed while maintaining less than 3 cm RMS positional error (Fig. \ref{fig:comparison}A). To perform sharp braking and turning, our robot could aggressively tilt its heading axis and quickly decelerate or accelerate. The maximum robot body deviation angle and lateral acceleration reached 50.0° and 11.7 m/s$^2$, representing 187\% and 254\% increase compared to prior results (Fig. \ref{fig:comparison}B). In addition, the robot could perform saccade movements under 160 cm/s wind disturbance – over 260\% that of a prior milestone \cite{chirarattananon2017dynamics}. It could further conduct ten consecutive body flips under the influence of unmodeled tension and weight from the power tether, which highlighted outstanding flight robustness absent in existing sub-gram robots. Our robot’s longest dimension measured 4 cm, yet it could reach a maximum flight speed of 197 cm/s, equivalent to 50 body lengths per second. This performance highlights the surprising result where soft-driven bio-inspired robots substantially outperformed rigid-driven systems at a similar scale.

These flight demonstrations exhibited similarities to that of aerial insects and were enabled by biomimetic hardware and controller designs. Fruit flies (\textit{Drosophila hydei}) generate large roll and pitch motions when they make evasive maneuvers in response to visual stimuli. Specifically, their body roll and pitch angles can increase by 45° within 40 ms \cite{muijres2014flies}, which requires a large change in thrust forces between consecutive wingbeats (5 ms). Our robot adopted a custom soft artificial muscle that drove the wing at 330 Hz – 70\% faster than that of flies. These actuators could quickly respond to electrical driving signals and produce large kinematics and thrust force changes within 3 ms. Consequently, our robot could achieve large normalized acceleration (F/mg) and aggressive pitching and rolling. In a body saccade maneuver, the normalized acceleration and body rotation of our robot and fruit flies \cite{muijres2014flies} exhibited similar magnitudes (Fig. \ref{fig:comparison}C). In addition, our robot's top flight speed and its ability to withstand wind disturbance were also comparable to that of flies \cite{leitch2021long}. 
These exceptionally fast body dynamics and control were enabled by bio-inspired actuators and mechanisms such as power-dense DEAs and flapping-wing propulsion. It would be infeasible to demonstrate similar levels of performance with conventional hardware such as microscale motors and rotors.

From a control perspective, our deep-learned RTMPC design addressed major challenges relating to agile insect flight: large model uncertainty, high susceptibility to environmental disturbances, and requirement on a high feedback rate. The tradeoff between controller complexity and computational cost was a major limiting factor for insect-scale aerial robots. Most prior work \cite{kim2025acrobatics,chen2019controlled,ma2013controlled,bena2023high,dhingra2025modeling} implemented proportional-derivative or state space controllers for executing low speed ($<$40 cm/s) trajectories at the expense of flight agility and robustness. To resolve this tradeoff, we demonstrated a two-step approach: a RTMPC for optimizing flight performance under large uncertainty followed by IL that generated a computationally efficient NN controller trained from the RTMPC policy.   

Compared to prior flight controllers, the RTMPC demonstrated substantial performance improvement when the robot executed aggressive maneuvers under unmodeled uncertainty and disturbance. For example, when the robot performed fast descent (Fig. \ref{fig:long_traj}C), it experienced 5\% lift force reduction due to unmodeled downwash effects. The tension and weight of the robot’s power tether also became substantial during body saccades (Fig. \ref{fig:controller}D) and flips (Fig. \ref{fig:flip}C). Our RTMPC handled these uncertainties and maintained the robot states within the tube. To execute the controller in real time, we designed a two-layer fully connected NN that mimics the central nervous system and the architecture of motor neurons found in fruit flies \cite{lesser2024synaptic}. Through IL, this NN is trained from the RTMPC and could run in real time on a desktop CPU. This controller design featured high performance and computational efficiency, leading to insect-like flight maneuvers that far exceed prior results \cite{kim2025acrobatics,chen2019controlled,ma2013controlled,bena2023high,dhingra2025modeling}. 

Despite showcasing insect-like flight capabilities, our robot relied on offboard motion sensing, compute, and power sources – a substantial gap compared to aerial insects. In flight experiments presented in Fig. \ref{fig:overview}-\ref{fig:flip}, the NN controller had 128 neurons in its intermediate layers and it updated at 1000 Hz. This computational requirement could be easily met with a desktop CPU but remained expensive for a resource-constrained micro controller unit (MCU) (e.g., the 10 mg and 120 MHz STM32F051). Since our controller design involved two separate stages, the compression step could be further optimized for onboard deployment. In both simulations and experiments, we investigated the influence of varying controller feedback rate and the number of hidden layer neurons on positional accuracy of body saccade maneuvers (Fig. \ref{fig:comparison}D-E). 

To maintain flight stability, aerial insects must sense and respond to their kinematic states at least once each wingbeat \cite{chang2014predicting}. In our flight setup, the motion capture system had a sensing rate of approximately 400 Hz – moderately above the 330 Hz flapping-wing frequency. We fixed the UKF update rate at 1000 Hz to ensure consistent state estimation quality under different controller feedback rates. The simulations and experiments shared a similar trend where RMS flight positional error and controller feedback rate were inversely correlated (Fig. \ref{fig:comparison}D). The robot flight performance and robustness experienced moderate changes when the feedback rate was above 200 Hz. In these cases, the RMS positional error changed within 119\% and all flights were successful. In contrast, flight performance deteriorated when the feedback rate was reduced to 50 Hz. We observed a 472\% increase of position error and the saccade maneuver success rate dropped by 60\%. These experiments showed the controller could tradeoff moderate flight performance for up to 10 times saving in computational cost.  

In addition, we further experimented with different NN sizes for commanding body saccade maneuvers. The computational cost of each controller update is proportional to the square of neuron number in hidden layers (See Supplementary Text for detailed analysis). Our simulations and experiments showed the robot could fly stably with a minimum of 8 neurons – over 92 times reduction of computation cost at the expense of 126\% increase in positional RMS error (Fig. \ref{fig:comparison}E). With an 8 neuron-sized NN, the controller could be easily executed on an onboard MCU weighing tens of milligrams. We envision future controllers will combine multiple NNs that create a library of motion primitives (e.g., saccade, ascent, descent, flip, etc). A complex flight trajectory could be composed of these primitives where the MCU would store several small NNs and select them based on high-level commands. The robot payload could also accommodate MEMs sensors such as time of flight, optical flow, and inertial measurement units. Our results on RTMPC and NN training open opportunities for future studies to pursue high performance insect-scale flights with onboard sensors and MCUs.     

\subsection*{MATERIALS AND METHODS}

\subsubsection*{Robot dynamic model}\label{sec:robot_dynamics}

The robot consists of four flapping-wing modules that operate at 330 Hz. The wing mass accounts for less than 2\% of the total robot mass, which implies that the robot can be modeled as a rigid body with states of position, velocity, attitude quaternion and angular velocity $\mathbf{x} = [\mathbf{p}^T, \mathbf{v}^T, \mathbf{q}^T, \boldsymbol{\omega}^T]^T$. Since the flapping-wing frequency is substantially faster than the robot body dynamics, the dynamic model only considers the time-averaged lift force produced by each flapping-wing module. The time-averaged drag force from each module is zero because the flapping-wing motion is reciprocal. As illustrated in fig. S1, the lift forces produced from the four modules create a net thrust force and torque with respect to the robot body. The thrust force is aligned with the robot body z-axis, and the torques act along the body x- and y-axes. Unlike quadcopter designs, this flapping-wing robot cannot generate yaw torque. The input command from the controller is given by $\mathbf{u}=[F_{\text{cmd}}, \tau_{x,\text{cmd}}, \tau_{y,\text{cmd}}]^T$. The system dynamics equations are formulated as     
\begin{align}\label{eq:robot_dynamics}
\dot{\mathbf{p}} &= \mathbf{v} \\
    \dot{\mathbf{v}} &= \frac{1}{m}(\mathbf{R} \; [0, 0, F_{\text{cmd}}]^T + [0, 0, -mg]^T + \Fdrag )\\
    \dot{\mathbf{q}} &= \frac{1}{2}\boldsymbol{\Lambda}(\boldsymbol{\omega})\cdot \mathbf{q} \\
    \boldsymbol{\dot{\omega}} &= \mathbf{I}^{-1}(-\boldsymbol{\omega} \times \! \mathbf{I} \boldsymbol{\omega} + [\tau_{x,\text{cmd}}, \tau_{y,\text{cmd}}, 0]^T + \taudrag) \label{eq:robot_dynamics_rot}
\end{align}
where $m$ is the robot mass, $\mathbf{R}$ is the rotation matrix from the robot frame to the world frame, $g$ is the gravity constant, $\mathbf{F}_{\text{drag}}$ is the drag force vector acting on the robot body in the world frame, $\boldsymbol{\Lambda}(\boldsymbol{\omega})$ is a skew-symmetric matrix generated by the angular velocity vector $\boldsymbol{\omega}$, $\mathbf{I}$ is the moment of inertia tensor, and $\boldsymbol{\tau}_{\text{drag}}$ is the rotational drag torque vector on the robot body in the body frame. 

\subsubsection*{Safe reference trajectory generator}
The reference trajectory consists of reference states, $\mathbf{X}^{\text{ref}} = [\mathbf{x}_0^{\text{ref}}, \mathbf{x}_1^{\text{ref}}, \ldots, \mathbf{x}_T^{\text{ref}}]$, and the reference control input, $\mathbf{U}^{\text{ref}} = [\mathbf{u}_0^{\text{ref}}, \mathbf{u}_1^{\text{ref}}, \ldots, \mathbf{u}_T^{\text{ref}}]$,  at each time step. The reference trajectory that is fed into nonlinear ancillary model predictive controller is either designed manually or obtained through trajectory optimization. For the first case, once a fourth order smooth sequence is designed, the corresponding control input sequence can be computed to satisfy the differential flatness property. For more aggressive trajectories, like somersaults, a trajectory optimization is executed under the system dynamics constraints to obtain the reference state and control input. 
\begin{align} 
\label{eq:ref_traj_gen}
    \Uref, \Xref
    = \underset{\mathbf{U}, \mathbf{X}}{\text{argmin}} \:\: &  
        \sum_{i=0}^{T} ( \;
            t_{\text{total}} + 
            \| \mathbf{x}_{i}\|^2_{\mathbf{Q}_{\text{ref}}} + \|\mathbf{u}_{i}\|^2_{\mathbf{R}_{\text{ref}}}\; ) \\
    \text{subject to} \:\: &  \mathbf{x}_{i+1} = f(\mathbf{x}_{i}, \mathbf{u}_{i})  \\ 
    & \mathbf{S}_k \mathbf{x}_{j} = \mathbf{x}_{\text{waypoint},k}, \quad \forall k = 1, \ldots, n \\
    & \mathbf{u}_{i} \in \mathbb{U}^{\text{ref}}, \quad \forall i = 0, \ldots , {T} \\
    & \mathbf{x}_{i} \in \mathbb{X}^{\text{ref}}, \quad \forall i = 0, \ldots , {T} 
\end{align}
where $j$ is the timestep at which waypoint $k$ must be met; $\mathbf{S}_k$ is a matrix to select the type of state variables to be constrained; \(\mathbf{Q}_{\text{ref}}\) is a positive-semi-definite weighting matrix and \(\mathbf{R}_{\text{ref}}\) is a positive-definite weighting matrix. In both cases, we ensure $\xref_j \in \Xsetref \subset \Xsetmpc \subseteq \mathbb{R}^{n_x} \; \forall j$, where $\mathbb{X}^{\text{ref}}  $ and represents the safe states the robot can operate in when accounting for uncertainties, and $\Xsetmpc$ the limit states if uncertainties were not present (e.g. position boundary or maximum velocity). We also enforce $\uref_j \in \Usetref \subset \Usetmpc \subseteq \mathbb{R}^{n_u} \; \forall j$, where  $\Usetmpc$ encodes the actuation limits of the robot hardware, and $\Usetref $ encodes tighter (more conservative) actuation limit that account for the effects of uncertainties.

\subsubsection*{Nonlinear ancillary model predictive control}
To track the reference trajectory, a \acf{NMPC} is implemented to maintain the state of the dynamical system close to the reference. This control problem is formulated as:
\begin{align} 
\label{eq:ancillary_nmpc_eq}
    \mathbf{U}_t^*, \mathbf{X}_t^*
    = \underset{\mathbf{U}, \mathbf{X}}{\text{argmin}} \:\: & 
        \| \mathbf{x}_{N} - \mathbf{x}^{\text{ref}}_{t+N} \|^2_{\mathbf{P}}  + 
        \sum_{i=0}^{N-1} ( \;
            \| \mathbf{e}_{i} \|^2_{\mathbf{Q}_{\text{mpc}}} + 
            \| \mathbf{u}_{i} - \mathbf{u}^{\text{ref}}_{t+i} \|^2_{\mathbf{R}_{\text{mpc}}} \; ) \\
    \text{subject to} \:\: &  \mathbf{x}_{i+1} = f(\mathbf{x}_{i}, \mathbf{u}_{i})  \\ 
    & \mathbf{x}_{0} = \mathbf{x}_{\text{current}} \\
    & \mathbf{u}_{i} \in \mathbb{U}^{\text{mpc}}, \forall i = 0, \ldots , {N-1} 
\end{align}
where \(\mathbf{Q}_{\text{mpc}}\), \(\mathbf{R}_{\text{mpc}}\), and \(\mathbf{P}\) are positive-definite weighting matrices. \(\mathbf{e}_{i}\) is the state tracking error at time step $i$. For position, velocity, and angular velocity, the planned and reference states are subtracted to compute the tracking error. For attitude, the 3-parameter Modified Rodrigues Parameter (MRP) \cite{shuster1993survey} attitude error representation is used. The index \(t\) marks the start of the reference window and increments as the robot advances. At every time step the problem is solved with the current state \(\mathbf{x}_{\text{current}}\); the horizon length is \(N\), which is set to 100 with a discretization time of 0.05 seconds in this work. The first element \(\mathbf{u}_{0}^{*}\) of \(\mathbf{U}_{t}^{*}\) is applied to the robot and appended to the optimal control sequence~\(\mathbf{U}^{\text{mpc}}\) as $\umpc_t$. More control margin is added to $\Usetmpc$ compared to the safe reference control sequence, $\Usetref$.

\subsubsection*{Tube construction}
As a component of the RTMPC implementation, we surround the nominal NMPC trajectory with a disturbance-invariant tube whose cross-sections \(\mathbb{T}^{x}\subset\mathbb{R}^{n_x}\) and \(\mathbb{T}^{u}\subset\mathbb{R}^{n_u}\) bound the state and the control input respectively at every time step. The tube is estimated through closed-loop Monte-Carlo simulations of the perturbed dynamics
\begin{equation} 
\mathbf{x}_{i+1}=f\!\bigl(\mathbf{x}_{i},\mathbf{u}_{i}\bigr)+\mathbf{w}_{i},\qquad
\mathbf{w}_{i}\in\mathbb{W},
\end{equation}
where \(\mathbb{W}\subset\mathbb{R}^{n_{x}}\) is a convex disturbance set that contains the origin. For each draw \(\mathbf{w}_{t}\in\mathbb{W}\) we propagate the closed-loop system and verify that the actual trajectory respects the constraint sets \(\Xsetmpc\) and \(\Usetmpc\), thereby confirming the feasibility of the reference plan. The result allows us to construct an axis-aligned hyper-rectangle, yielding time-invariant bounds \(\mathbb{T}^{x}\) for the state and \(\mathbb{T}^{u}\) for the input. By construction, any initial state \(\mathbf{x}_{0}\in \Xmpc\) together with any disturbance sequence in \(\mathbb{W}\) remains inside \(\mathbb{T}^{x}\); an analogous statement holds for \(\mathbb{T}^{u}\). This tube, derived from the nominal NMPC trajectory and the constant deviation vector \(\Delta\mathbf{x}\), not only certifies that all disturbed trajectories stay within the bounds but also defines the sampling domain used later for data augmentation in IL.


\subsubsection*{Data augmentation via parametric sensitivities} \label{subsec:sensitivity}

The tube \(\mathbb{T}^{x}\) produced by the NMPC in Eq.\ref{eq:ancillary_nmpc_eq} bounds the region of state space that the robot can reach under disturbances. This region conveniently defines the domain for training the NN controller via \acf{DA}. Generating additional state–action pairs by re-solving NMPC for every sampled state is impractical, as each call requires solving a sequential quadratic program (SQP). To reduce this cost we apply \acf{SA}, which approximates the NMPC locally with a time-varying linear model. At each time step \(t\) we form the auxiliary set \(\mathcal{X}_{t}=\{\mathbf{x}_{t},\,t,\,\mathbf{X}^{\text{ref}},\,\mathbf{U}^{\text{ref}}\}\) and compute the first-order sensitivity matrix of the optimal input \(\umpc_{t}\) with respect to the state.
\begin{equation} \label{eq:sensitivity_matrix}
\mathbf{K}_{\mathcal{X}_t} \, \coloneqq \,
\left. \frac{\partial \umpc_t}{\partial \mathbf{x}_t} 
\right |_{\mathcal{X}_t }
\,  = \,
\begin{bmatrix}
\left. \dfrac{\partial \umpc_t}{\partial{ x_{1_t}}}
\right |_{\mathcal{X}_t},\,& 
\,
\dots \, , 
\,
& \,
\left. \dfrac{\partial \umpc_t}{\partial{ x_{n_{x_t}}}}
\right |_{\mathcal{X}_t} & 
\\
\end{bmatrix}.
\end{equation}
The sensitivity matrix $\mathbf{K}_{\mathcal{X}_t} \in \mathbb{R}^{n_u \times n_x}$, whose columns contain the directional derivatives of the control inputs with respect to the states, also called the tangential predictor, enables us to compute extra actions $\uaug_{t,k}$ by sampling states inside the tube $\xaug_{t,k} \in \mathbb{T}^x$, with $k=1,\ldots,N_k$: 
\begin{equation}
\label{eq:approximate_ancillary_controller}
    \uaug_{t,k} = \umpc_{t} + \mathbf{K}_{\mathcal{X}_{t}} (\xaug_{t,k} - \xmpc_{t}) 
\end{equation}
The proposed approximation makes the data-augmentation approach computationally efficient as a full SQP need not be solved for every state-action pairs \((\xaug_{t,k},\uaug_{t,k})\). Instead, the sensitivity matrix \(\mathbf{K}_{\mathcal{X}_t}\) is evaluated once per time step and reused for all samples drawn from the tube at that instant. Linearization points in~\cref{eq:sensitivity_matrix} are taken along the actual trajectory acquired during the \ac{NMPC} demonstration collection. These \ac{NMPC} demonstrations rely on the complete SQP solution, whereas the sensitivity-based linear surrogate is employed exclusively for the subsequent data-augmentation phase.

\subsubsection*{Imitation learning of neural network controller}
We train a computationally efficient NN policy that replicates the expert \acs{RTMPC} using IL. The chosen network architecture is a multi-layer perceptron with two fully connected hidden layers of equal size (8-128 neurons each). The training dataset for a given trajectory is generated from expert demonstrations collected in simulation and efficiently augmented using the sensitivity-based sampling method described above. More specifically, data collection and policy training are performed across an initial phase and a fine-tuning phase, using few demonstrations. First, in the initial phase, a single demonstration is collected to carry out Behavior Cloning, and the resulting training dataset is augmented with 100–200 samples per timestep. Given a simulation timestep of 0.005 s over a 5–10 s trajectory, this initial dataset contains 100,000–400,000 state-action pairs. An initial NN policy is trained using the mean squared error loss and the Adam optimizer \cite{kingma2014adamICLR} for 300–600 epochs, with early-stopping based on validation loss, using a patience of 30-60 epochs. Then, in the fine-tuning phase, the training dataset generated in the initial phase is discarded, and additional fine-tuning demonstrations are collected sequentially, typically for a total of 10 demonstrations, to update the NN policy after each new demonstration. This fine-tuning phase uses the same training loss as the initial phase but requires significantly fewer epochs, as it operates on a smaller dataset without data augmentation. The intermediate policy after each demonstration is stored, and the final NN control policy that will be deployed to the robot is determined through evaluation under disturbances in a high-fidelity simulation environment. We found that deploying the policies trained after 3-6 demonstrations yields the best performance on the robot hardware.

\subsubsection*{Unscented Kalman Filter}

Because the motion capture system only tracks position and orientation, we design an \ac{UKF}~\cite{wan2000unscented,tagliabue2019robust,tagliabue2017collaborative,simon2006optimal} to estimate full robot states including position, attitude, translational velocity, angular velocity, and the unknown external torque~$\boldsymbol{\tau}_{\text{ext}}$.  
With the disturbance torque included, the rotational dynamic equation becomes
\begin{equation} 
\boldsymbol{\dot{\omega}} = \mathbf{I}^{-1}(-\boldsymbol{\omega} \times \mathbf{I} \boldsymbol{\omega} + [\tau_{x,\text{cmd}}, \tau_{y,\text{cmd}}, 0]^\top + \boldsymbol{\tau}_{\text{drag}} + \boldsymbol{\tau}_{\text{ext}}),
\label{eq:external_torque_dynamics}
\end{equation}
and the new augmented state becomes
\begin{equation}
\mathbf s = \begin{bmatrix} \mathbf p^\top, \; \mathbf q^\top, \; \mathbf v^\top, \; \boldsymbol{\omega}^\top, \; \boldsymbol{\tau}_{\text{ext}}^\top \end{bmatrix}^\top \in \mathbb{R}^{16}.
\label{eq:stateVectorsOneLine}
\end{equation}
To define the rest of the filter, we introduce an alternative representation $\mathcal{X}$ of the state, in which the quaternion attitude is replaced with the three-parameter Modified Rodrigues Parameter (MRP) ~$\mathbf{e}$:
\begin{equation}
\mathcal{X} = \begin{bmatrix} \mathbf p^\top, \; \mathbf e^\top, \; \mathbf v^\top, \; \boldsymbol{\omega}^\top, \; \boldsymbol{\tau}_{\text{ext}}^\top \end{bmatrix}^\top \in \mathbb{R}^{15}.
\end{equation}
Note that we use MRPs to represent the difference between a particular attitude and a reference attitude \cite{tagliabue2017collaborative}.
Given a reference attitude $\mathbf{q}_\text{ref}$, one can convert from attitude $\mathbf{q}$ to MRP $\mathbf{e}$ and vice versa.
We also define \( \hat{\mathbf{s}}_k^{-} \equiv \hat{\mathbf{s}}_{k|k-1} \) to be the estimated state before measurement at time step $k$, and \( \hat{\mathbf{s}}_k^{+} \equiv \hat{\mathbf{s}}_{k|k} \) to be the estimated state after incorporating measurement.


The filter consists of two steps, a prediction step that uses the dynamics model to compute $\hat{\mathbf{s}}^-_{k}$ given $\hat{\mathbf{s}}^+_{k-1}$ and an update step that incorporates measurement $\mathbf{z}_k$ into $\hat{\mathbf{s}}^-_{k}$ to obtain $\hat{\mathbf{s}}^+_{k}$.
To perform the prediction step, we first convert $\hat{\mathbf{s}}_{k-1}^+$ into the alternative representation, using the identity MRP $\mathbf{0}_{3\times1}^\top$ to represent no difference from our reference attitude of $\hat{\mathbf{q}}_{k-1}^+$. Then, we 
generate sigma points using the covariance matrix from the prior time step:
\begin{align} \label{eq:sigma_points}
\mathcal{X}_{k-1}^0 &=
\begin{bmatrix}
\hat{\mathbf{p}}^\top_{k-1}, \;
\mathbf{0}_{3\times1}^\top, \;
\hat{\mathbf{v}}^\top_{k-1}, \;
\hat{\boldsymbol{\omega}}^\top_{k-1}, \;
\hat{\boldsymbol{\tau}}_{\text{ext},{k-1}}^\top
\end{bmatrix}^\top,\\
\mathcal{X}_{k-1}^l = \mathcal{X}_{k-1}^0 + &\left[\sqrt{(n + \lambda)\, \mathbf{P}^+_{k-1}}\,\right]_l, \qquad
\mathcal{X}_{k-1}^{l+n} = \mathcal{X}_{k-1}^0 - \left[\sqrt{(n + \lambda)\, \mathbf{P}^+_{k-1}}\,\right]_l,
\end{align}
for \( l = 1, \dots, n \), where \( n \) is the state dimension and \( \lambda \) is the UKF scaling parameter.  
The matrix square root is computed via Cholesky decomposition, and \([\,\cdot\,]_l\) denotes the \(l\)-th column. Using $\hat{\mathbf{q}}^+_{k-1}$ as reference, each sigma point $\mathcal{X}_{k-1}^l$ is converted back to the standard state representation $\mathbf{s}_{k-1}^l$ and is propagated through the dynamics:
\begin{equation}
\mathbf{s}_k^l = f_{\text{ukf}}(\mathbf{s}_{k-1}^l, \mathbf{u}_k), \quad l = 0, \dots, 2n.
\end{equation}
where \( f_{\text{ukf}} (\cdot) \) represents the discrete dynamics in which Eq. \ref{eq:external_torque_dynamics} replaces Eq. \ref{eq:robot_dynamics_rot} in Sec. \ref{sec:robot_dynamics}. 
The propagated sigma points $\mathbf{s}_k^l$ are then converted back to the MRP attitude state representation with respect to $\mathbf{q}^0_{k}$ to obtain $\mathcal{X}_{k}^l$, and are combined in a weighted average with UKF mean weights $w_m$ to obtain:
\begin{equation}
\hat{\mathcal{X}}^-_k = \sum_{l=0}^{2n} w_{m,l}\, \mathcal{X}_k^l.
\end{equation}
Finally, $\hat{\mathcal{X}}^-_k$ is converted to $\hat{\mathbf{s}}^-_k$ using $\mathbf{q}_k^0$.
The predicted covariance is
\begin{equation}
\mathbf{P}^-_k = \sum_{l=0}^{2n} w_{c,l} \bigl(\mathcal{X}_k^l - \hat{\mathcal{X}}^-_k\bigr)\bigl(\mathcal{X}_{k}^l - \hat{\mathcal{X}}^-_k\bigr)^\top + \mathbf{Q}_{\text{ukf}},
\end{equation}
where \( w_c \) are the UKF covariance weights.

Because the observation model is linear, the update step reduces to the classical Kalman filter form. The motion tracking system provides position~$\mathbf p_k^{m}$ and attitude~$\mathbf q_k^{m}$. A six-element measurement vector $\mathbf{z}_k$ is produced, with MRP $\mathbf{e}_k^m$ computed with respect to $\hat{\mathbf{q}}_k^-$:
\begin{equation}
\mathbf z_k = \begin{bmatrix} \mathbf p_k^{m} \\ \mathbf{e}_k^{m} \end{bmatrix} = \mathbf H \mathbf s_k + \mathbf v_k, \qquad
\mathbf H = \begin{bmatrix} \mathbf I_{6\times6} & \mathbf 0_{6\times9} \end{bmatrix},
\label{eq:measModelOneLine}
\end{equation}
and with measurement noise \(\mathbf v_k \sim \mathcal{N}(\mathbf 0, \mathbf R_{\text{ukf}})\). The Kalman gain is computed as
\begin{equation}
\mathbf K_k = \mathbf P_k^- \mathbf H^\top \bigl( \mathbf H \mathbf P_k^- \mathbf H^\top + \mathbf R_{\text{ukf}} \bigr)^{-1},
\label{eq:UKF_gain}
\end{equation}
which weighs the relative trust in the model prediction versus the new measurement.  
The covariance is updated via the Joseph stabilized covariance measurement update equation \cite{simon2006optimal}
\begin{equation}
\mathbf P_k^+ =
(\mathbf I - \mathbf K_k \mathbf H)\, \mathbf P_k^-\, (\mathbf I - \mathbf K_k \mathbf H)^\top + \mathbf{K}_k \mathbf{R}_\text{ukf} \mathbf{K}_k^\top
\label{eq:UKF_cov}
\end{equation}
and the state is updated as
\begin{equation}
\hat{\mathcal X}_k^+ = \hat{\mathcal X}_k^- + \mathbf K_k \bigl( \mathbf z_k - \mathbf H \hat{\mathbf s}_k^- \bigr),
\label{eq:UKF_state}
\end{equation}
and then converted to $\hat{\mathbf{s}}_k^+$ using $\hat{\mathbf{q}}_k^-$ as reference.

In summary, Eqs. \ref{eq:sigma_points}--\ref{eq:UKF_state} describe each UKF loop iteration, which estimates the full robot states (position, attitude, translational velocity, and angular velocity) and external torque based on position and attitude measurements.

\clearpage 

%
\bibliography{science-bib} 

%
%
%
%
%
%


\subsection*{Acknowledgments}

\noindent\textbf{Funding:} \\
\hspace*{2em} National Science Foundation FRR-2202477 (YC) \\
\hspace*{2em} National Science Foundation FRR-2236708 (YC) \\
\hspace*{2em} Office of Naval Research N00014-25-1-2303 (YC) \\
\hspace*{2em} Air Force Office of Scientific Research MURI FA9550-19-1-0386 (JPH) \\ 
\hspace*{2em} MathWorks Engineering Fellowship (YHH) \\
\hspace*{2em} MathWorks Research Equipment Grant (YC) \\
\hspace*{2em} Zakhartchenko Fellowship (SK) \\

\noindent\textbf{Author contributions:} \\
\hspace*{2em} Conceptualization: YHH, AT, JPH, YC \\
\hspace*{2em} Methodology: YHH, AT, OM, SK, ZT, JPH, YC \\
\hspace*{2em} Software: AT, YHH, OM, TZ \\
\hspace*{2em} Validation: YHH, AT, OM, SK \\
\hspace*{2em} Formal analysis: YHH, AT, OM, YC \\
\hspace*{2em} Investigation:  YHH, AT, OM, YC \\
\hspace*{2em} Visualization:  YHH, OM, SK \\
\hspace*{2em} Funding acquisition: YC, JPH \\
\hspace*{2em} Project administration: YC, JPH \\ 
\hspace*{2em} Supervision: YC, JPH \\
\hspace*{2em} Writing – original draft: YHH, OM, YC \\
\hspace*{2em} Writing – review \& editing: YHH, AT, OM, TZ, JPH, YC 

\paragraph*{Competing interests:}
The authors declare no competing interests.

\paragraph*{Data and materials availability:}
All data are available in the main text or the supplementary materials.


\subsection*{Supplementary materials}

\hspace*{2em} Supplementary Text\\
\hspace*{2em} Figs. S1 to S16\\
\hspace*{2em} Table S1\\
\hspace*{2em} Movies S1 to S7\\


\end{document}



\renewcommand{\thefigure}{S\arabic{figure}}
\renewcommand{\thetable}{S\arabic{table}}
\renewcommand{\theequation}{S\arabic{equation}}
\renewcommand{\thepage}{S\arabic{page}}
\setcounter{figure}{0}
\setcounter{table}{0}
\setcounter{equation}{0}
\setcounter{page}{1} 


\begin{center}
\vspace*{1.5cm}
\section*{Supplementary Materials for}

\Large \scititle

    \vspace*{0.5 cm}
    \normalsize
	Yi-Hsuan Hsiao$^{1\dagger}$,
	Andrea Tagliabue$^{2\dagger}$,
	Owen Matteson$^{2\dagger}$,
    Suhan Kim$^{1}$, 
    Tong Zhao$^{1,2}$,\\
    Jonathan P. How$^{2\ast}$,
    YuFeng Chen$^{1\ast}$ \\

    \begingroup
    \singlespacing
    \small
    $^{1}$Department of Electrical Engineering and Computer Science, Massachusetts Institute of Technology, \\
    Cambridge, Massachusetts, 02139, United States.\\
    $^{2}$Department of Aeronautics and Astronautics, Massachusetts Institute of Technology, \\
    Cambridge, Massachusetts, 02139, United States.\\

    \endgroup
    
	\small$^\ast$Corresponding authors. Emails: jhow@mit.edu, yufengc@mit.edu \\
	\small$^\dagger$These authors contributed equally to this work.
\end{center}

\textbf{This PDF file includes:} \\
\hspace*{4em} Supplementary Text\\
\hspace*{4em} Figs. S1 to S16\\
\hspace*{4em} Table S1\\
\hspace*{4em} Captions for Movies S1 to S7\\

\textbf{Other Supplementary Materials for this manuscript:} \\
\hspace*{4em} Movies S1 to S7\\

\newpage

\onehalfspacing 


\subsection*{Supplementary Text}

\vspace{1cm}

\subsubsection*{Flight experimental setup} \label{supsec:setup}
\vspace{5mm}

The motion capture system consists of six cameras (Vantage V5, Vicon), which provide real-time measurements of the robot’s position and orientation. This state information is transmitted to the control computer via asynchronous \ac{UDP}. The control computer runs an \ac{UKF} to estimate translational and angular velocities, which are then passed to the \ac{NN} controller. The controller computes the desired force for each of the four actuators, which is converted into corresponding voltage amplitudes, \(V_{\text{amp}}\), through a force-to-voltage mapping. These amplitudes are used to generate the actuator driving signals, shaped as square-root-modulated sine waves, \(V(t) = V_{\text{amp}} \frac{1}{\sqrt{2}}\sqrt{\sin(2\pi f t) + 1}\), where $f$ is the flapping frequency (330 Hz in this work). The resulting digital signals are sent to a \ac{DAC}, which converts them into \(0\text{–}10\,\mathrm{V}\) analog signals. These analog signals are then amplified by four high-voltage amplifiers (2220, Trek) to \(0\text{–}2000\,\mathrm{V}\) and delivered to the robot through five 49-AWG power tethers (49 QP 155 RED, MWS), comprising four high-voltage lines and one ground line. The robot is equipped with seven 1.5-mm motion-tracking markers (B\&L Engineering) for position and attitude tracking. \\

\noindent
The sampling frequency of the motion capture system slightly deviates from 400\,Hz due to occasional communication delay from the motion capture camera to the motion capture computer. To address this issue, the motion-tracking computer runs the \ac{UDP} communication script at a much higher rate and transmits the latest available data immediately upon receiving an update from the tracking system. The new state information is sent to the \ac{UKF}, which operates at 1000\,Hz to accommodate the asynchronous input and generate state estimates at 1000\,Hz for the \ac{NN} controller. The \ac{NN} controller also runs at 1000\,Hz with the exception of experiments explicitly testing different feedback rates. The voltage amplitude \(V_{\text{amp}}\), used to construct the driving signals, is updated at the beginning of the rising phase of each flapping cycle at 330\,Hz, matching the flapping frequency. The high-rate execution of the \ac{UKF} is implemented due to the non-uniform update intervals of the motion capture system and the mismatch between the sensing and actuation frequencies.

\newpage

\subsubsection*{Frame alignment, unit force allocation, and voltage mapping} \label{supsec:frame_alignment}
\vspace{5mm}
Our flapping-wing robot is unable to generate yaw torque and therefore lacks yaw control authority. Although the \ac{NN} policy is trained at a nominal yaw angle, the robot can enter yaw configurations during flight that are not represented in the training dataset. Instead of expanding the dataset to include all possible yaw angles, we implement a frame alignment procedure that constructs a correction rotation matrix, $\mathbf{R}_{\text{match}}$, to align the robot's current yaw angle with that used during \ac{NN} training. As a result, the rotational states passed to the \ac{NN} controller frame are defined as

\setcounter{equation}{0}

\begin{align}
    \mathbf{q}_{\text{nn}} &= \text{quat}(\mathbf{R}_{\text{match}}\mathbf{R})\\
    \boldsymbol{\omega}_{\text{nn}} &= \mathbf{R}_{\text{match}}\boldsymbol{\omega}
\end{align}
and the torque output from the \ac{NN}, computed in the \ac{NN} controller frame, is transformed back to the actual body frame via
\begin{align}
    \boldsymbol{\tau} = \mathbf{R}_{\text{match}}^T\boldsymbol{\tau}_{\text{nn}}
\end{align}

\noindent
The alignment matrix $\mathbf{R}_{\text{match}}$ is constructed as follows. First, the current body z-axis, given by the third column of the rotation matrix $\mathbf{R}_3$, is copied into the z-axis of the \ac{NN} controller frame:
\begin{align}
    \mathbf{R}_{3,\text{match}} = \mathbf{R}_3.
\end{align}
Next, the y-axis of the \ac{NN} frame is computed as the normalized cross product between the current z-axis and the desired body x-axis, represented by the first column of the reference rotation matrix $\mathbf{R}_1^{\text{mpc}}$ extracted from the \ac{MPC} trajectory $\Xmpc$:
\begin{align}
    \mathbf{R}_{2,\text{match}} = \frac{\mathbf{R}_3 \times \mathbf{R}_1^{\text{mpc}}}{||\mathbf{R}_3 \times \mathbf{R}_1^{\text{mpc}}||}
\end{align}
Finally, the x-axis is computed as the normalized cross product of the new y- and z-axes:
\begin{align}
    \mathbf{R}_{1,\text{match}} = \frac{\mathbf{R}_{2,\text{match}} \times \mathbf{R}_{3,\text{match}}}{||\mathbf{R}_{2,\text{match}} \times \mathbf{R}_{3,\text{match}}||}.
\end{align}
The complete alignment matrix is given by $\mathbf{R}_{\text{match}} = [\mathbf{R}_{1,\text{match}}, \; \mathbf{R}_{2,\text{match}}, \; \mathbf{R}_{3,\text{match}}]$. The reference x-axis $\mathbf{R}_1^{\text{mpc}}$ is time-varying and pre-computed from the trajectory sequence $\Xmpc$.\\

\noindent
Due to the lack of yaw torque generation, we implement a custom unit force allocation strategy based on the commanded total thrust \( \Fcmd \), commanded body torques about the x- and y-axes \( \taux \), \( \tauy \), and the estimated external torques \( \tau_{x,\text{ext}} \) and \( \tau_{y,\text{ext}} \). We define:
\begin{equation}
\mathbf{u}' 
=
\frac{1}{4}
\begin{bmatrix}
\frac{\taux - \tau_{x,\text{ext}}} {l_x} \\
\frac{\tauy - \tau_{y,\text{ext}}} {l_y}
\end{bmatrix}
\end{equation}
where \( l_x \) and \( l_y \) are the respective moment arms about the x- and y-axes. The allocation matrix is defined as:
\begin{equation}
\mathbf{A}' = 
\begin{bmatrix}
-1 & -1 \\
\;\;\,1 & -1 \\
\;\;\,1 & \;\;\,1 \\
-1 & \;\;\,1 \\
\end{bmatrix}
\end{equation}
The thrust differentials required to achieve the desired torques are given by:
\begin{equation}
\boldsymbol{\delta} = \mathbf{A}' \mathbf{u}' 
\end{equation}
Assuming an equal nominal thrust across all actuators, we define the initial unit force as \( f_{ini} = \Fcmd / 4 \). To ensure all unit forces remain non-negative, we adjust this value as:
\begin{equation}
f_{ini}' \leftarrow \max\left( f_{ini} + \min_i \delta_i \;\;, \;\;0\right)
\end{equation}
where \( \delta_i \) denotes the \(i\)-th element of the vector \( \boldsymbol{\delta} \). The final unit force for each actuator is then computed as:
\begin{equation}
f_i = f_{ini}' + \delta_i  \;\;\;\; \forall i = 1, 2, 3, 4
\end{equation}
This allocation strategy ensures torque tracking along the x- and y-axes of the body frame while maintaining non-negativity of actuator thrusts. Its closed-form and low computational cost make it well-suited for real-time implementation on resource-constrained systems. \\

\noindent
The computed unit forces are subsequently converted into voltage amplitudes using an empirical force-to-voltage mapping, which is experimentally characterized for each individual actuator. This mapping accounts for hardware-specific variations and ensures accurate voltage generation to achieve the desired thrust output. The relationship is expressed as:
\begin{equation}\label{eq:force_to_voltage}
    V_{\text{amp},i} = \text{f2v}_{i}(f_i) \;\;\;\; \forall i = 1, 2, 3, 4.
\end{equation}
Here, \(\text{f2v}_{i}(\cdot)\) denotes the calibrated mapping function for the \(i\)-th actuator, translating the desired force \(f_i\) into the corresponding voltage amplitude \(V_{\text{amp},i}\) required to drive the actuator appropriately.

\newpage

\subsubsection*{Simulation}
\vspace{5mm}
The robotic simulation used in this manuscript follows the dynamics described by the equations below:
\begin{align}\label{eq:robot_sim_dynamics}
\dot{\mathbf{p}} &= \mathbf{v} \\
\dot{\mathbf{v}} &= \frac{1}{m}(\mathbf{R} \; [0, 0, F_{\text{cmd}}']^T + [0, 0, -mg]^T + \Fdrag + \mathbf{F}_\text{ext} )\\
\dot{\mathbf{q}} &= \frac{1}{2}\boldsymbol{\Lambda}(\boldsymbol{\omega})\cdot \mathbf{q} \\
\boldsymbol{\dot{\omega}} &= \mathbf{I}^{-1}(-\boldsymbol{\omega} \times \! \mathbf{I} \boldsymbol{\omega} + [\tau_{x,\text{cmd}}', \tau_{y,\text{cmd}}', 0]^T + \taudrag + \boldsymbol{\tau}_\text{ext}). \label{eq:robot_sim_omega_dynamics}
\end{align}

\noindent
Compared to the robot dynamics model outlined in the Methods section, the simulation introduces external force disturbances \(\mathbf{F}_\text{ext}\) and torque disturbances \(\boldsymbol{\tau}_\text{ext}\) to perturb the robot away from its nominal trajectory. These additions make the simulation more representative of real-world experimental conditions. The control inputs are denoted as \(F_{\text{cmd}}'\), \(\tau_{x,\text{cmd}}'\), and \(\tau_{y,\text{cmd}}'\), which differ from the direct outputs of the \ac{NN} controller. The prime notation (\('\)) indicates that these inputs incorporate actuator saturation, communication delays, and second-order dynamics arising from the flapping-wing actuation mechanism. These modified inputs are obtained as follows. \\

\noindent
First, the voltage amplitude of each individual actuator is saturated:
\begin{equation}
    V_{\text{amp},i}^{\text{sat}} = \min ( V_{\text{amp},i} \;, \;V_{\text{max},i} ), \;\;\;\; \forall i = 1, 2, 3, 4
\end{equation}
where \(V_{\text{max},i}\) represents the maximal safe operating voltage amplitude for the \(i\)-th actuator, beyond which electrical breakdown may occur. The saturated voltage is then passed through the inverse of the force-to-voltage function defined in Eq.~\ref{eq:force_to_voltage}. (Note that Eq.~\ref{eq:force_to_voltage} is monotonic within the actuator’s operational range. If increasing voltage yields no additional thrust, the voltage is capped at the point where the derivative vanishes: \(\{V_{\text{max},i} \in \mathbb{R} \; | \; (\text{f2v}_i^{-1})'(V_{\text{amp},i})=0\}\). This ensures that \(\text{f2v}_i(\cdot)\) remains invertible.) The saturated unit force is then computed as:
\begin{equation}
    f_{i}^{\text{sat}} = \text{f2v}_i^{-1} ( V_{\text{amp},i}^{\text{sat}}), \;\;\;\; \forall i = 1, 2, 3, 4
\end{equation}

\noindent
To account for the actuation dynamics of the flapping-wing system, we apply a second-order low-pass Butterworth filter to the saturated unit forces:
\begin{equation}
    \ddot{f}_i^{\text{fil}} + \sqrt{2}\,\omega_c \dot{f}_i^{\text{fil}} + \omega_c^2 f_i^{\text{fil}} = \omega_c^2 f_i^{\text{sat}}, \;\;\;\; \forall i = 1, 2, 3, 4
\end{equation}
where \(\omega_c\) is the cutoff frequency of the filter. Next, to simulate communication delay between the sensing cameras, sensing computer, control computer, and \ac{DAC}, we apply a time delay to the filtered unit force:
\begin{equation}
    f_i'(t) = f^{\text{fil}}_i(t-t_d), \;\;\;\; \forall i = 1, 2, 3, 4
\end{equation}

\noindent
With the delayed unit forces \(f_i'\), we compute the collective thrust and body torques using the standard allocation matrix:
\begin{equation}
\begin{bmatrix}
\Fcmd' \\
\taux' \\
\tauy'
\end{bmatrix}
=
\begin{bmatrix}
1 & 1 & 1 & 1 \\
-l_x & l_x & l_x & -l_x \\
-l_y & -l_y & l_y & l_y
\end{bmatrix}
\begin{bmatrix}
f_1' \\
f_2' \\
f_3' \\
f_4'
\end{bmatrix}
\end{equation}

\noindent
External disturbances \(\mathbf{F}_\text{ext}\) and \(\boldsymbol{\tau}_\text{ext}\) are generated using a random walk process. The modified control inputs \(F_{\text{cmd}}'\), \(\tau_{x,\text{cmd}}'\), and \(\tau_{y,\text{cmd}}'\) are then applied in the dynamic equations given in Eq. \ref{eq:robot_sim_dynamics}-\ref{eq:robot_sim_omega_dynamics}. The simulation is implemented in Matlab Simulink, where the continuous-time state evolution is integrated using Simulink's built-in ODE solver.

\newpage

\subsubsection*{Tube safety strategy}
\vspace{5mm}
The robust tube not only defines a bounded region within which the expert controller can safely operate, but also characterizes the domain over which the \acf{NN} policy is trained. Ensuring that the system state remains within this tube during execution is critical for maintaining policy safety and avoiding extrapolation into unseen regions. This section describes the real-time strategy used to verify that the current state remains within the tube, as well as the mechanism designed to ensure a safe landing in the event of a significant deviation.\\

\noindent
During flight, the controller continuously checks whether the current state lies within the predefined tube. To accommodate variations in yaw—which are not explicitly handled by the neural network policy—an alternative representation is used for the tube safety check. Instead of using quaternions for orientation, we compute the angles between the robot’s body-frame $z$-axis and each of the world-frame axes:
\(\boldsymbol{\theta} = [ \arccos(R_{13}) \; \arccos(R_{23}) \; \arccos(R_{33}) ] ^T\)
For angular velocity, we use the realigned vector $\boldsymbol{\omega}_{\text{nn}}$, as described in the previous sections. With these substitutions, we define the alternative state vector as
\(\mathbf{x}' = [ \mathbf{p}^T, \: \mathbf{v}^T , \: \boldsymbol{\theta}^T , \: \boldsymbol{\omega}_{\text{nn}}^T ]^T. \)
This formulation allows the tube cross-section $\Delta\mathbf{x}'$ to be applied directly in a hyper-rectangular form. To verify whether the robot remains within the tube, we evaluate whether \( \mathbf{x}' \in \left[ \mathbf{x}^{'\text{mpc}} - \Delta\mathbf{x}',\; \mathbf{x}^{'\text{mpc}} + \Delta\mathbf{x}' \right]. \) \\

\noindent
If the state remains outside the tube for a duration exceeding a predefined threshold, the control architecture transitions from the learned policy to a model-based controller. Given the high angular velocities that may arise from aggressive maneuvers, a geometric attitude controller is first employed to dampen rotational motion and restore an upright orientation, without attempting positional stabilization. Once the angular velocity falls below a safe threshold, a position controller is activated to bring the robot to a stable, controlled landing. This hierarchical strategy enables the system to pursue aggressive performance using the \acf{NN} controller, while still responding gracefully to large deviations or unmodelled disturbances to maintain safety and flight integrity.

\newpage

\subsubsection*{Computational cost of neural network}
\vspace{5mm}
The neural networks employed in this work are multi-layer perceptrons consisting of two fully connected hidden layers of equal size, with neuron counts ranging from 8 to 128. The primary computational burden arises from matrix multiplications during forward propagation. Given 14 input features from the \ac{UKF}, the input layer requires \(14 \times n\) operations, where \(n\) is the number of neurons per hidden layer. Each hidden layer involves \(n \times n\) operations, and the output layer (with three outputs) requires an additional \(3 \times n\) operations. Therefore, the total computational complexity scales as \(n^2 + 17n\). For instance, a network with 128 neurons incurs approximately 92.8 times the computational cost of a network with only 8 neurons. \\

\newpage


\begin{figure}[H]
\centering
\includegraphics[width=180mm]{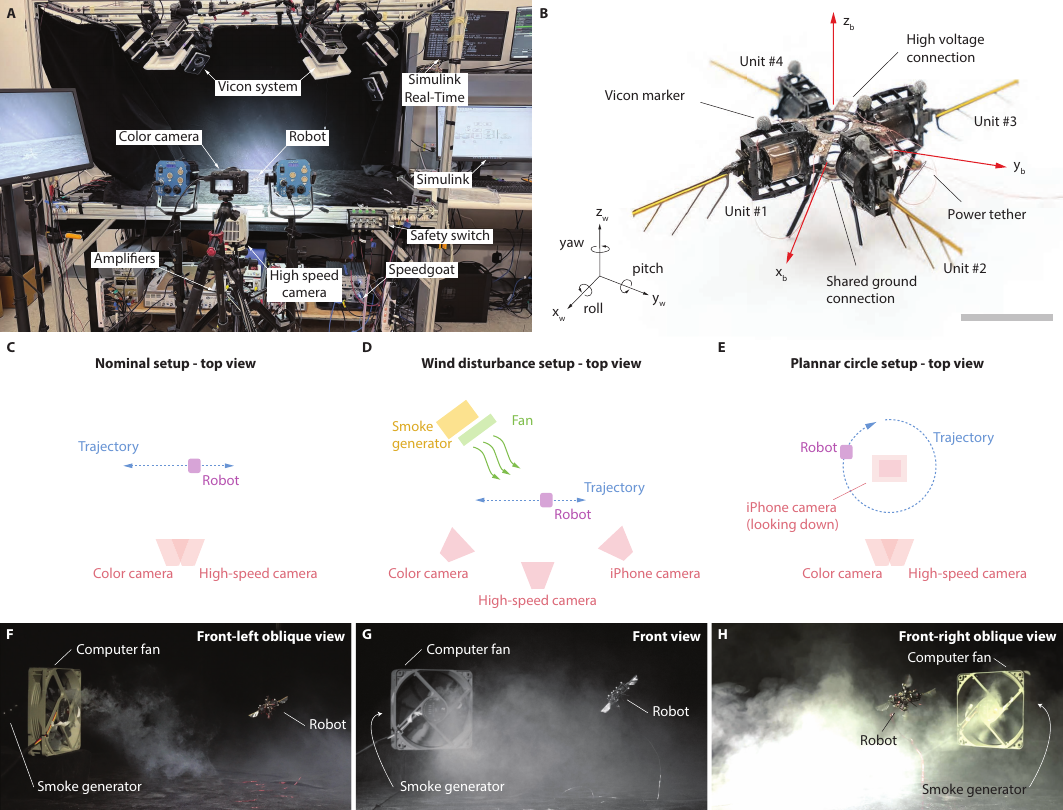}
\vspace*{-0.5cm}
\caption{\textbf{Experimental setup for flight demonstrations.} (\textbf{A}) Our flight arena consists of a Vicon motion capture system, a custom Speedgoat control computer that runs the Simulink Real-Time controller, high-voltage amplifiers, and a set of cameras. (\textbf{B}) Coordinate system definition for constructing the flight controller. (\textbf{C}) Nominal flight experimental setup for maneuvers including the body saccade, repeated saccade, body flips, "X"-shaped, "+"-shaped, and figure 8 trajectories. (\textbf{D}) Experimental setup for disturbance rejection saccade maneuvers. Three cameras were placed at different perspectives. (\textbf{E}) Experimental setup for filming the circular flight trajectory along the xy-plane. (\textbf{F}-\textbf{H}) Sample images of the three perspectives corresponding to (D).}\label{figsup:setup}
\end{figure}
\newpage

\begin{figure}[H]
\centering
\includegraphics[width=180mm]{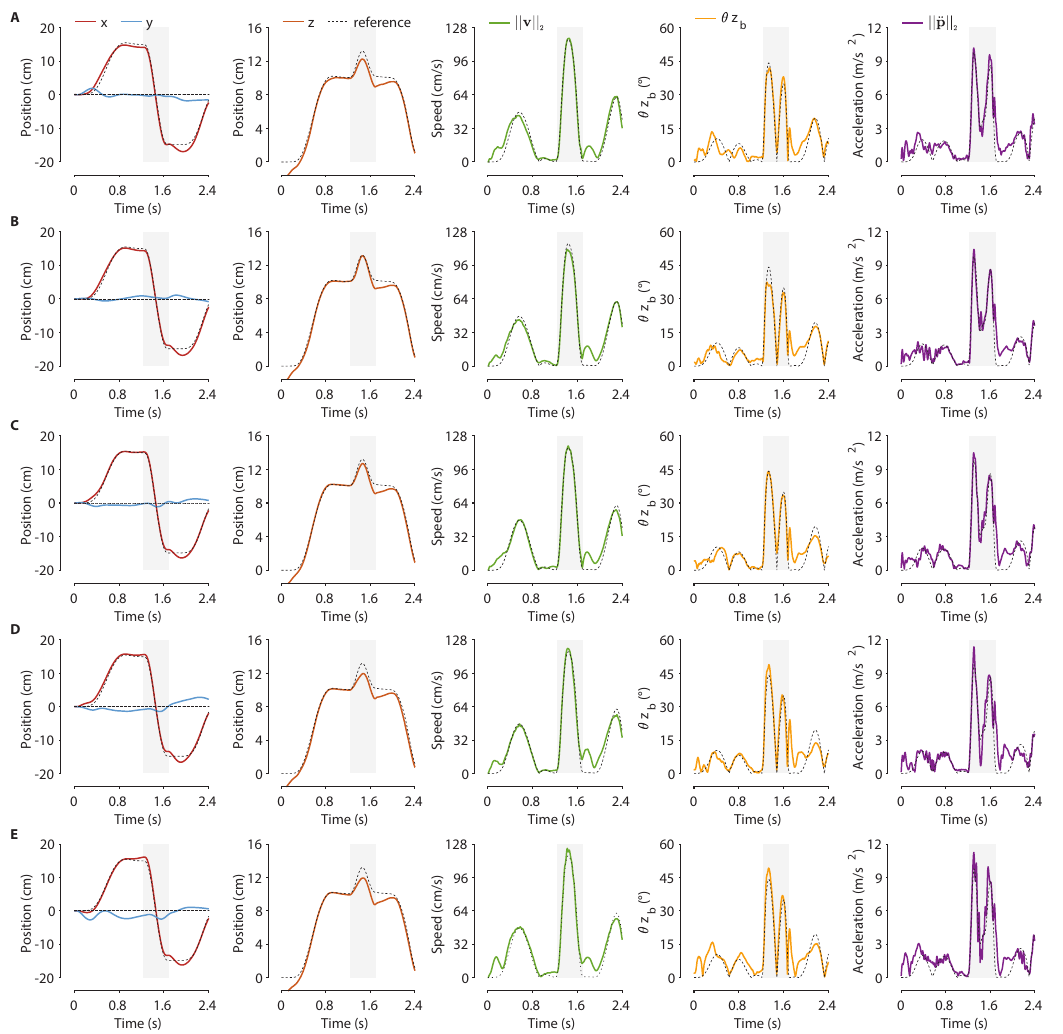}
\vspace*{-0.5cm}
\caption{\textbf{Five point-to-point flight demonstrations.} Each horizontal panel (\textbf{A}-\textbf{E}) shows data from one experiment. The five columns show x and y position, altitude, speed, body deviation angle, and acceleration. The shaded region highlights the dynamic maneuver during which flight metrics (speed, positional error, etc) are computed from. }\label{figsup:p2p}
\end{figure}
\newpage

\begin{figure}[H]
\centering
\includegraphics[width=180mm]{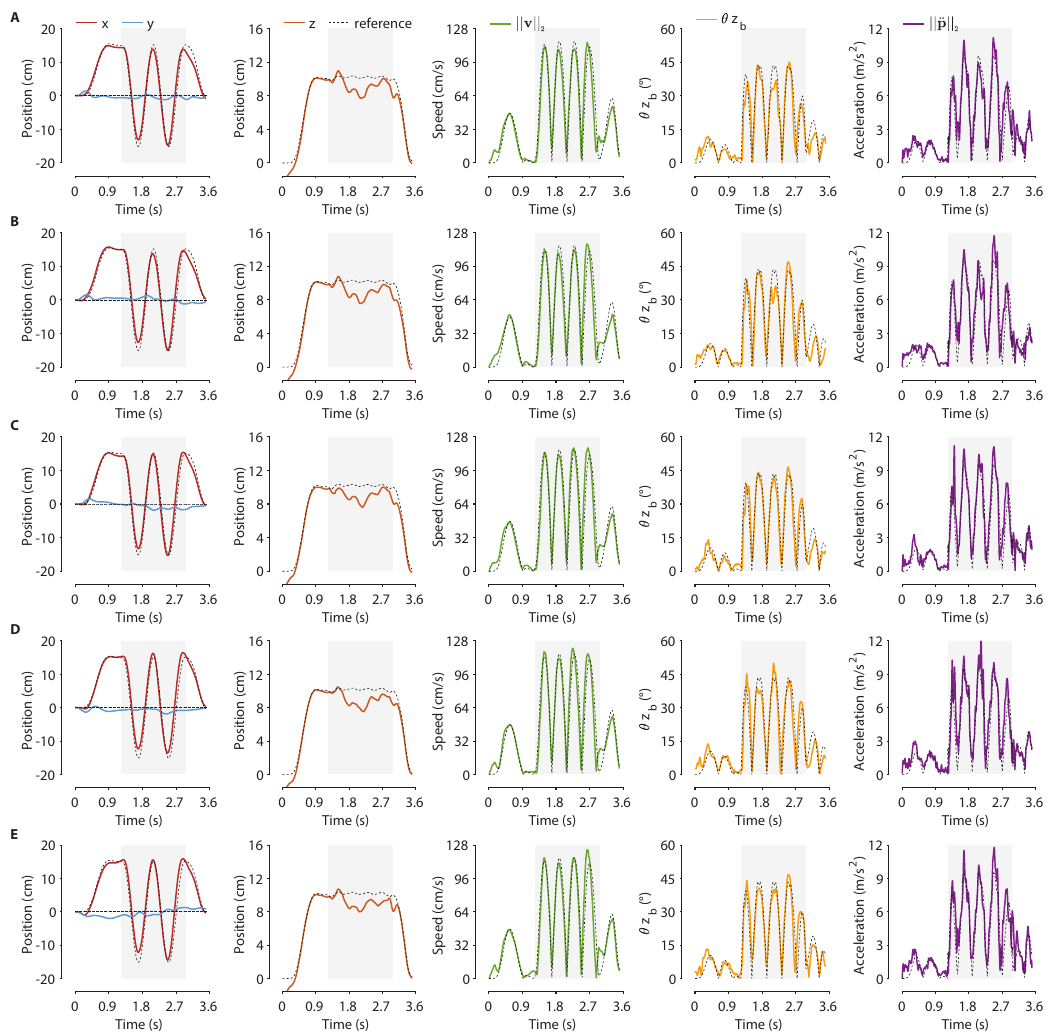}
\vspace*{-0.5cm}
\caption{\textbf{Five multi-point-to-point flight demonstrations.} Each horizontal panel (\textbf{A}-\textbf{E}) shows data from one experiment. The five columns show x and y position, altitude, speed, body deviation angle, and acceleration. The shaded region highlights the dynamic maneuver during which flight metrics (speed, positional error, etc) are computed from.  }\label{figsup:multi_p2p}
\end{figure}
\newpage

\begin{figure}[H]
\centering
\includegraphics[width=180mm]{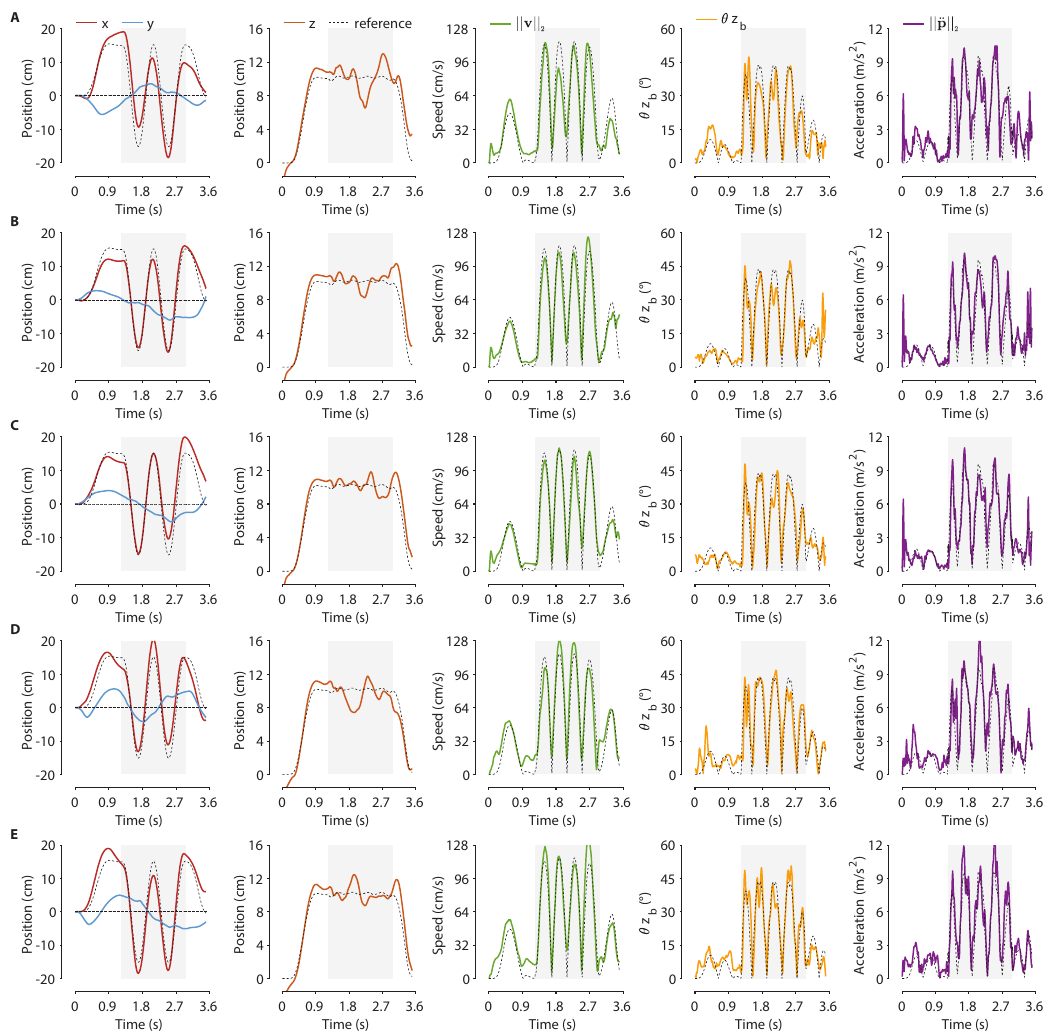}
\vspace*{-0.5cm}
\caption{\textbf{Five multi-point-to-point flight demonstrations under 33\% force-to-voltage mapping error.} Each horizontal panel (\textbf{A}-\textbf{E}) shows data from one experiment. The five columns show x and y position, altitude, speed, body deviation angle, and acceleration. The shaded region highlights the dynamic maneuver during which flight metrics (speed, positional error, etc) are computed from.  }\label{figsup:multi_p2p_incorrect}
\end{figure}
\newpage

\begin{figure}[H]
\centering
\includegraphics[width=180mm]{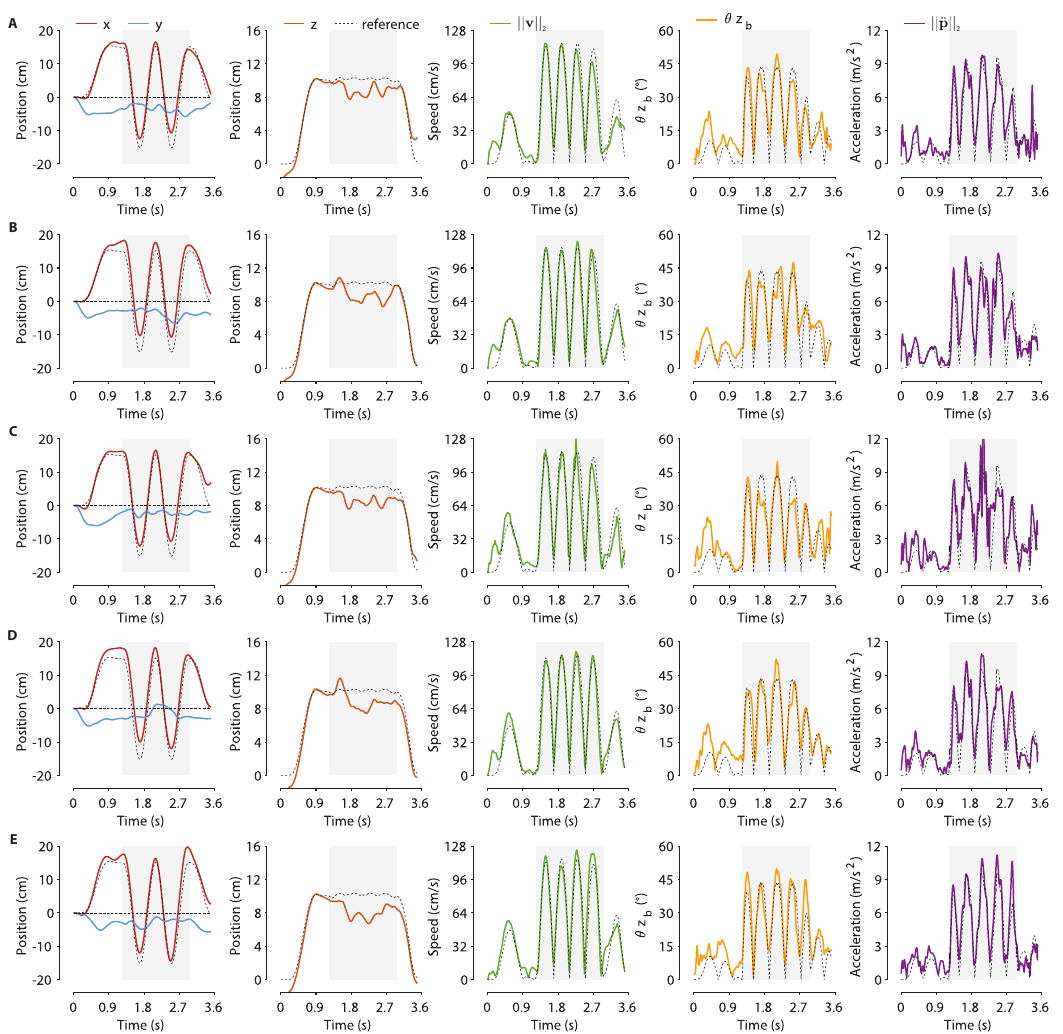}
\vspace*{-0.5cm}
\caption{\textbf{Five multi-point-to-point flight demonstrations under wind disturbance.} Each horizontal panel (\textbf{A}-\textbf{E}) shows data from one experiment. The five columns show x and y position, altitude, speed, body deviation angle, and acceleration. The shaded region highlights the dynamic maneuver during which flight metrics (speed, positional error, etc) are computed from.  }\label{figsup:multi_p2p_disturb}
\end{figure}
\newpage

\begin{figure}[H]
\centering
\includegraphics[width=180mm]{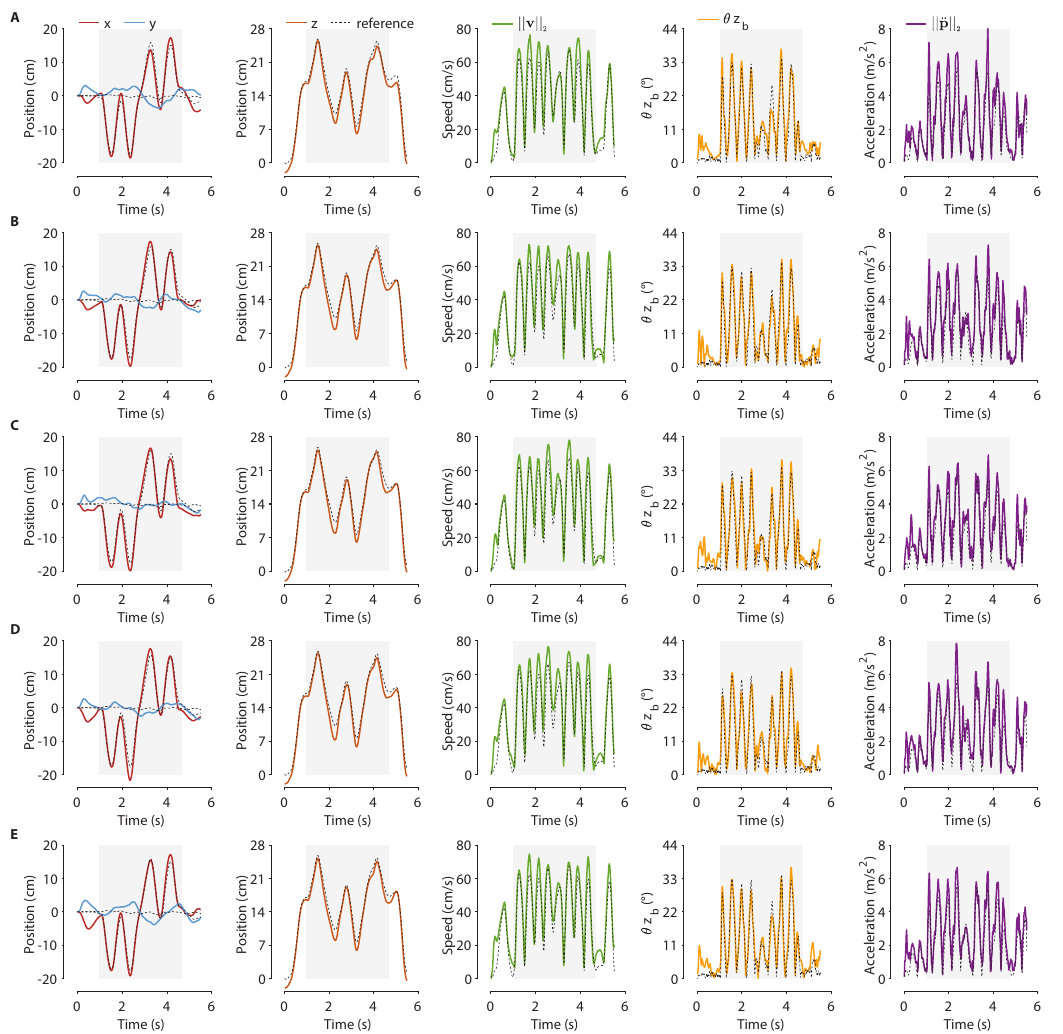}
\vspace*{-0.5cm}
\caption{\textbf{Five X-shaped trajectory tracking demonstrations.} Each horizontal panel (\textbf{A}-\textbf{E}) shows data from one experiment. The five columns show x and y position, altitude, speed, body deviation angle, and acceleration. The shaded region highlights the dynamic maneuver during which flight metrics (speed, positional error, etc) are computed from.  }\label{figsup:x_shape}
\end{figure}
\newpage

\begin{figure}[H]
\centering
\includegraphics[width=180mm]{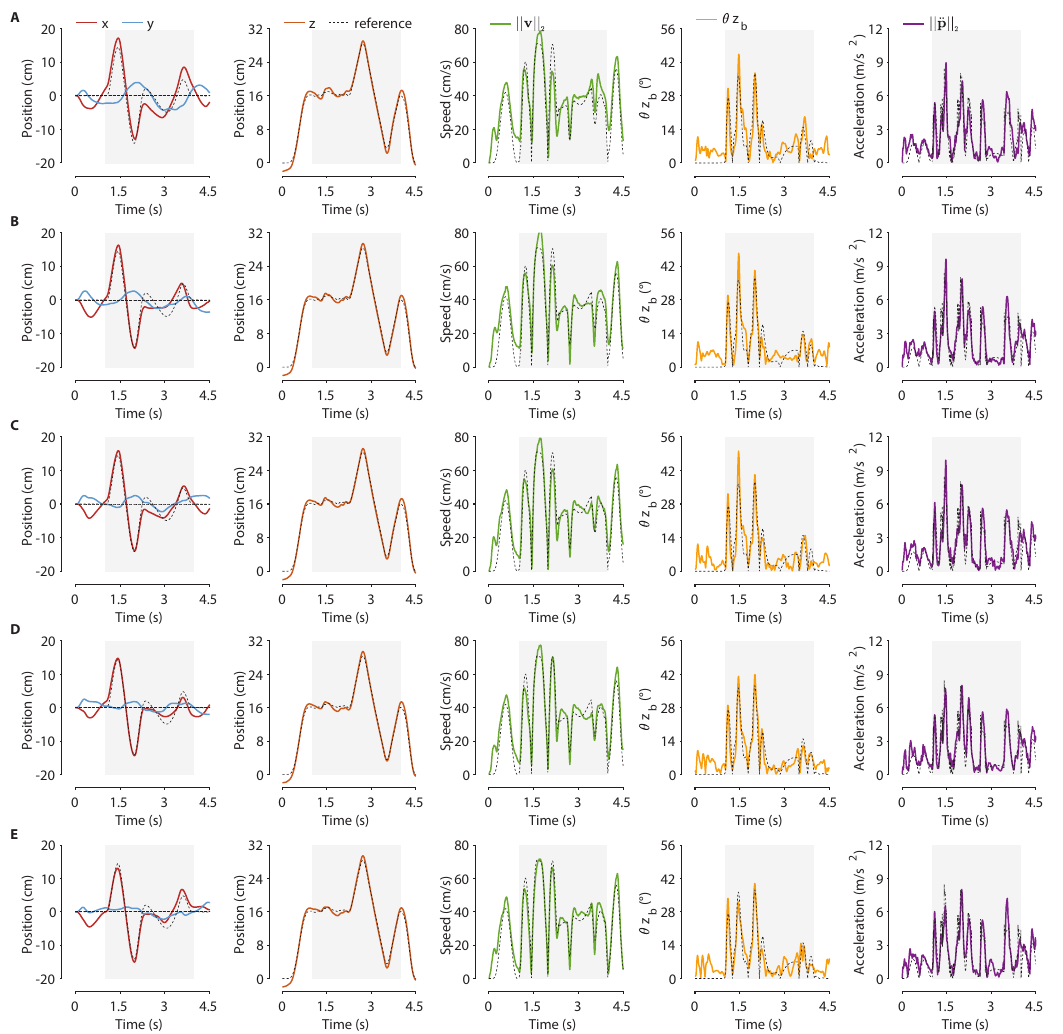}
\vspace*{-0.5cm}
\caption{\textbf{Five ``+"-shaped trajectory tracking demonstrations.} Each horizontal panel (\textbf{A}-\textbf{E}) shows data from one experiment. The five columns show x and y position, altitude, speed, body deviation angle, and acceleration. The shaded region highlights the dynamic maneuver during which flight metrics (speed, positional error, etc) are computed from.  }\label{figsup:plus_sign}
\end{figure}
\newpage

\begin{figure}[H]
\centering
\includegraphics[width=180mm]{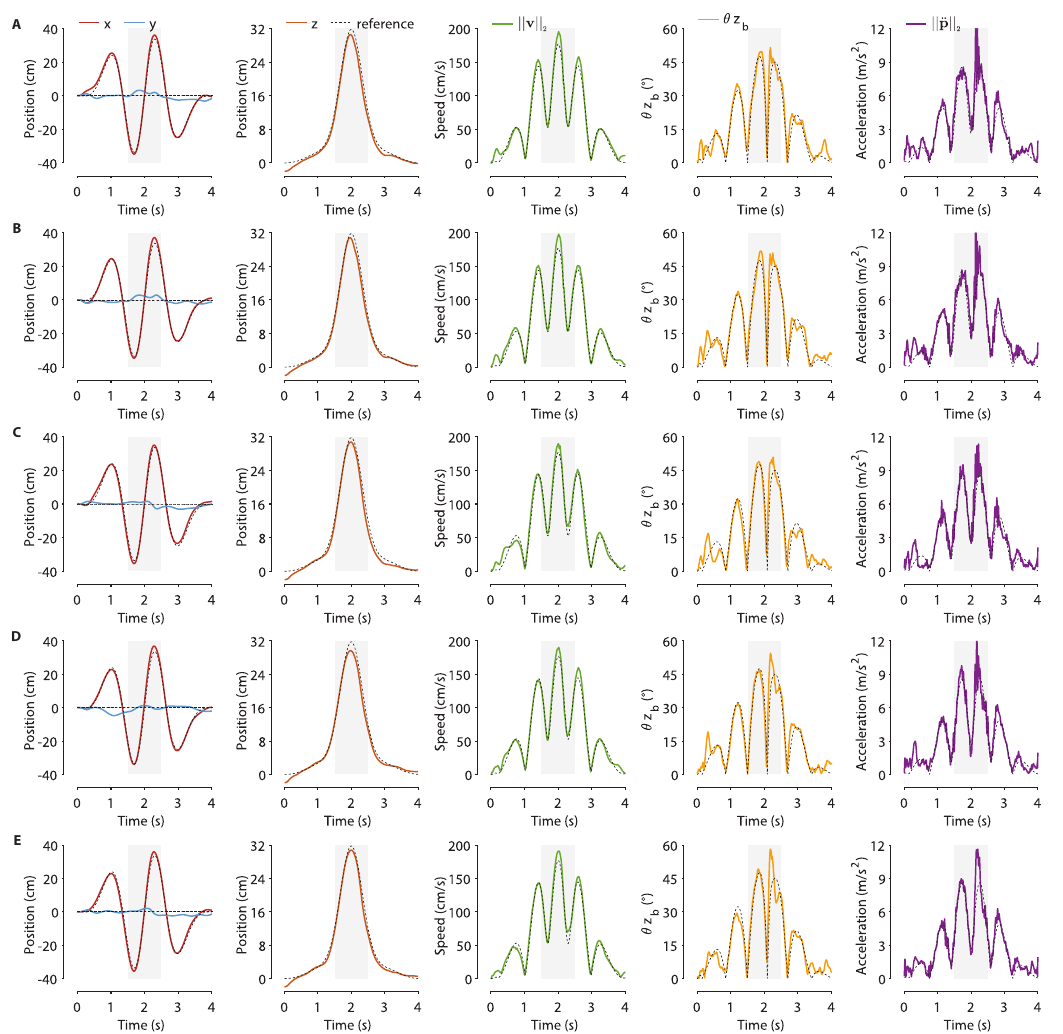}
\vspace*{-0.5cm}
\caption{\textbf{Five figure 8 trajectory tracking demonstrations.} Each horizontal panel (\textbf{A}-\textbf{E}) shows data from one experiment. The five columns show x and y position, altitude, speed, body deviation angle, and acceleration. The shaded region highlights the dynamic maneuver during which flight metrics (speed, positional error, etc) are computed from.  }\label{figsup:fast_vertical}
\end{figure}
\newpage

\begin{figure}[H]
\centering
\includegraphics[width=180mm]{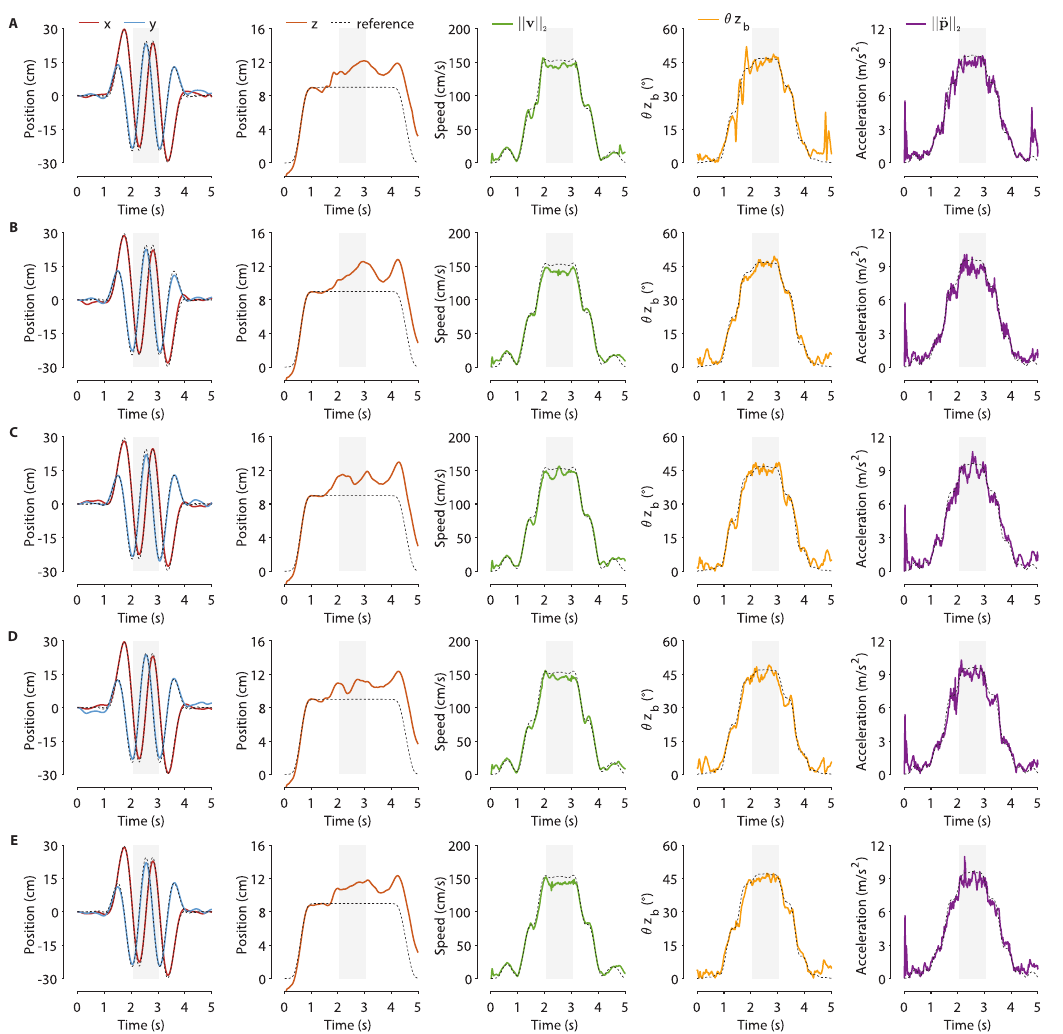}
\vspace*{-0.5cm}
\caption{\textbf{Five planar circle tracking demonstrations.} Each horizontal panel (\textbf{A}-\textbf{E}) shows data from one experiment. The five columns show x and y position, altitude, speed, body deviation angle, and acceleration. The shaded region highlights the dynamic maneuver during which flight metrics (speed, positional error, etc) are computed from. }\label{figsup:fast_horizontal}
\end{figure}
\newpage

\begin{figure}[H]
\centering
\includegraphics[width=180mm]{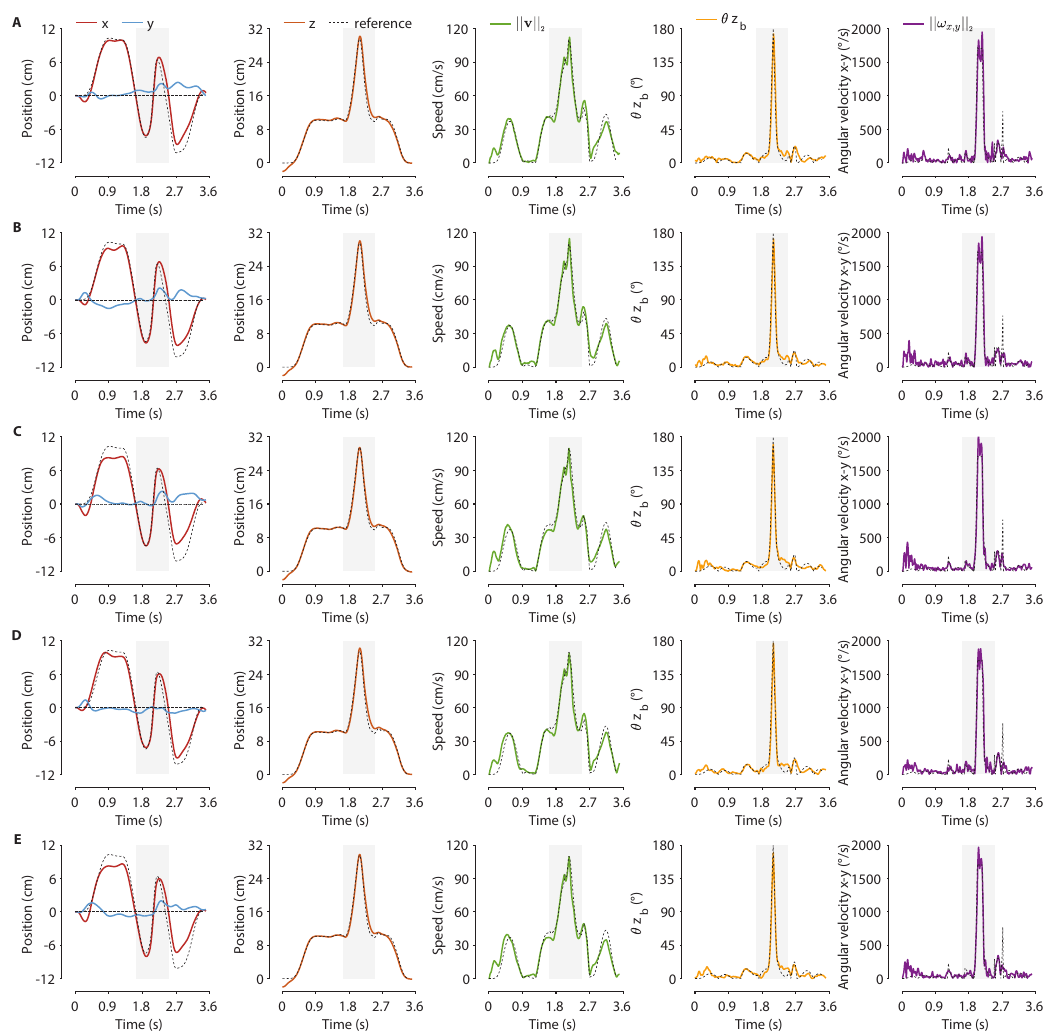}
\vspace*{-0.5cm}
\caption{\textbf{Five single body flip demonstrations.} Each horizontal panel (\textbf{A}-\textbf{E}) shows data from one experiment. The five columns show x and y position, altitude, speed, body deviation angle, and angular speed. The shaded region highlights the dynamic maneuver during which flight metrics (speed, positional error, etc) are computed from. }\label{figsup:flip_single}
\end{figure}
\newpage

\begin{figure}[H]
\centering
\includegraphics[width=180mm]{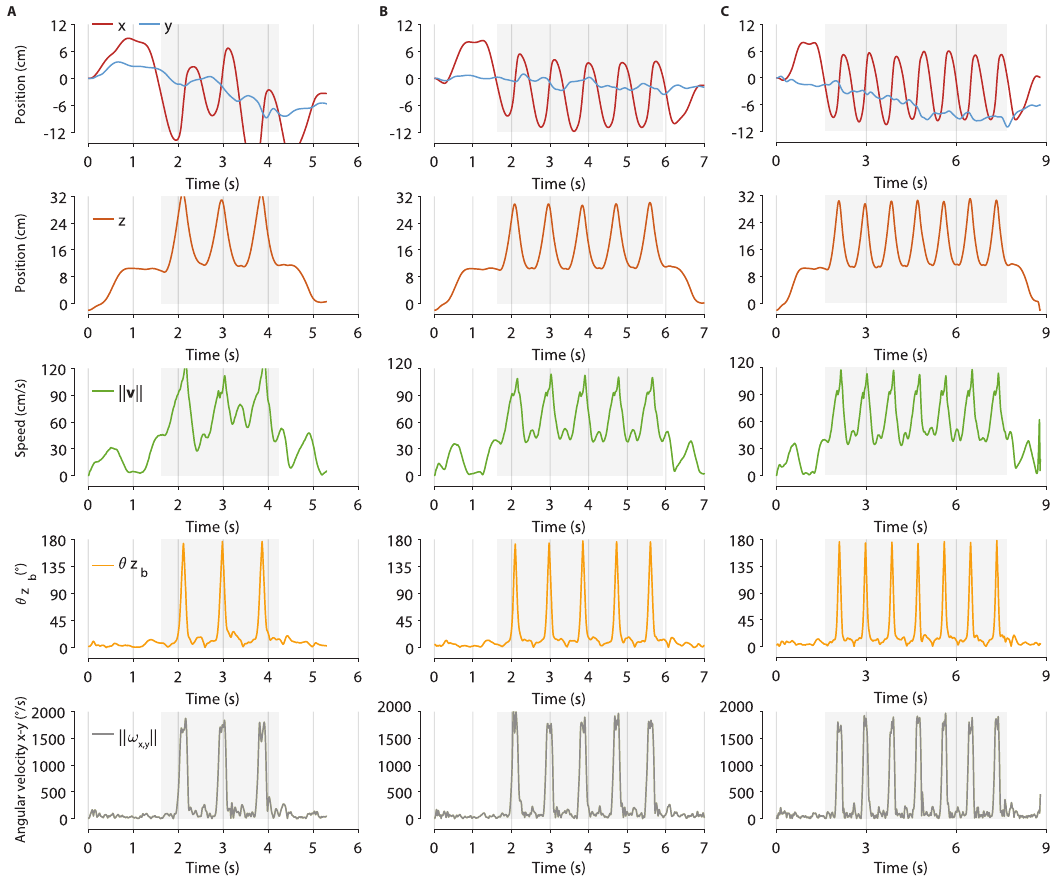}
\vspace*{-0.5cm}
\caption{\textbf{Consecutive body flip demonstrations.} Each vertical column (\textbf{A}-\textbf{C}) shows data from one experiment. The five rows show x and y position, altitude, speed, body deviation angle, and angular speed. The shaded region highlights the dynamic maneuver during which flight metrics (speed, positional error, etc) are computed from. (A-C) shows three, five, and seven consecutive body flips, respectively. }\label{figsup:flip_multi}
\end{figure}
\newpage

\begin{figure}[H]
\centering
\includegraphics[width=180mm]{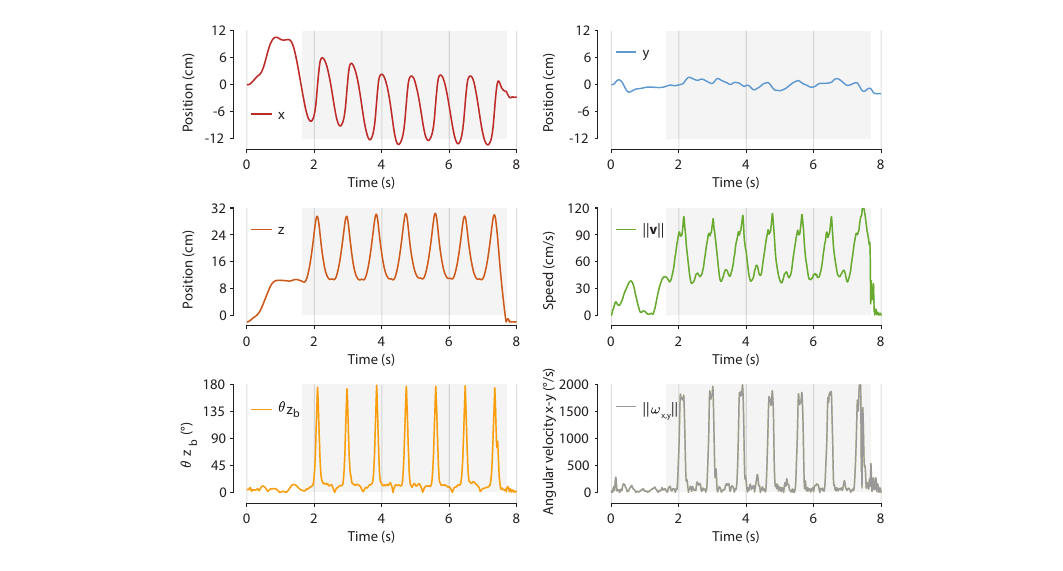}
\vspace*{-0.5cm}
\caption{\textbf{Aborted body flip demonstration.} When the robot states deviated outside the tube in a body flip experiment, a model-based controller was switched on to stabilize robot attitude and land. The panels show tracked position, altitude, speed, body deviation angle, and angular speed. $\theta_{z_b}$ remained small when the flight was aborted and the robot landed on the ground with a upright orientation, preventing the damage of the fragile wings.}\label{figsup:flip_aborted}
\end{figure}
\newpage

\begin{figure}[H]
\centering
\includegraphics[width=180mm]{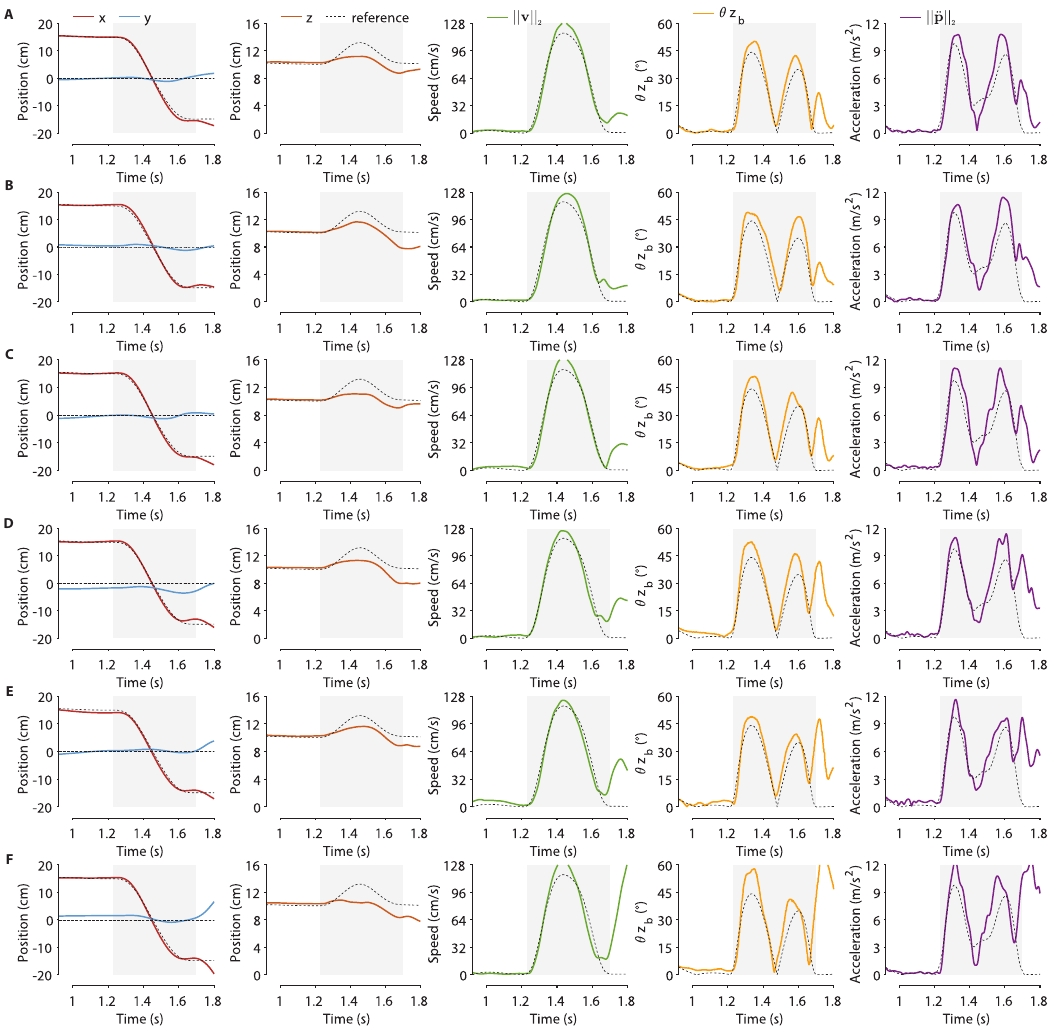}
\vspace*{-0.5cm}
\caption{\textbf{Point-to-point flight demonstrations under 500 Hz and 330 Hz feedback rate.} Each horizontal panel (\textbf{A}-\textbf{F}) shows data from one experiment. The five columns show x and y position, altitude, speed, body deviation angle, and acceleration. The shaded region highlights the dynamic maneuver during which flight metrics (speed, positional error, etc) are computed from. (A-C) and (D-F) correspond to 500 Hz and 330 Hz feedback rates, respectively.   }\label{figsup:multi_rate_500_330}
\end{figure}
\newpage

\begin{figure}[H]
\centering
\includegraphics[width=180mm]{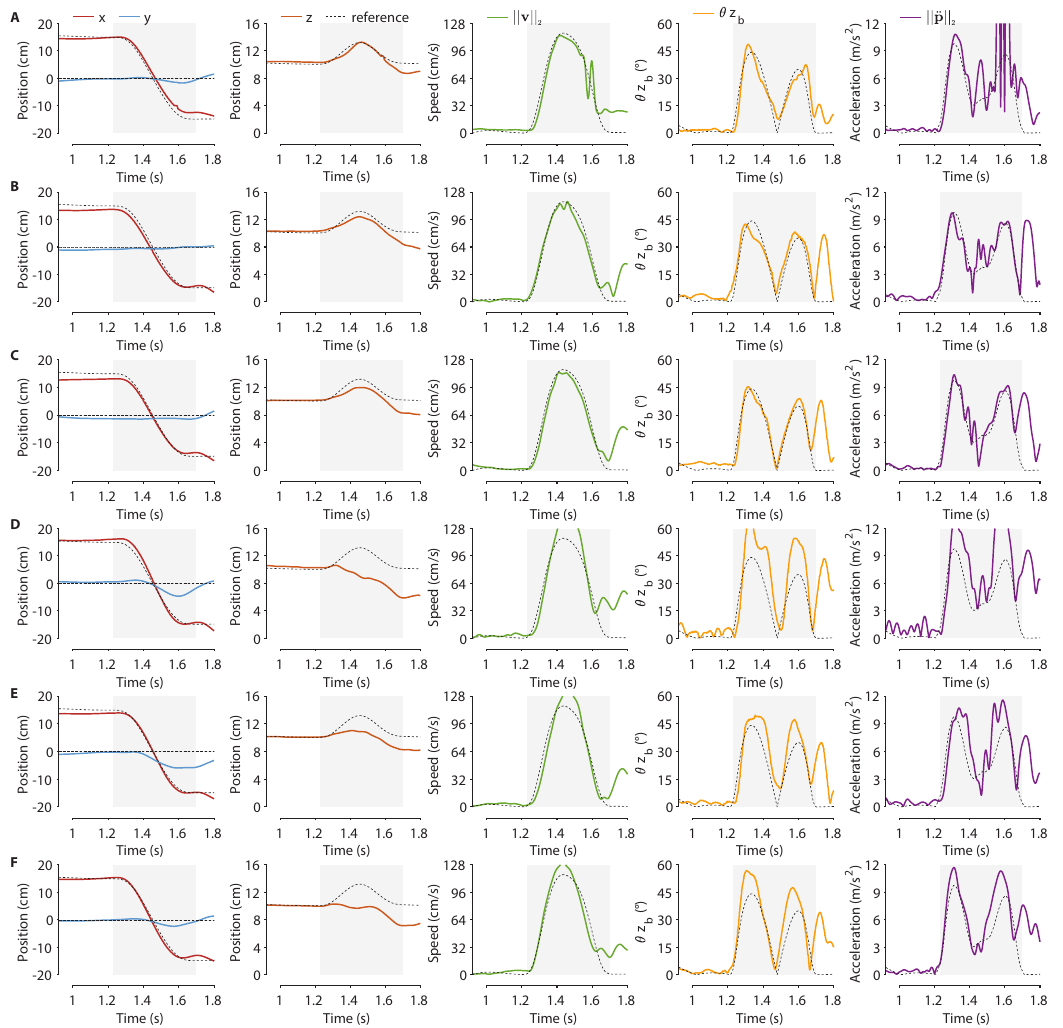}
\vspace*{-0.5cm}
\caption{\textbf{Point-to-point flight demonstrations under 200 Hz and 100 Hz feedback rate.} Each horizontal panel (\textbf{A}-\textbf{F}) shows data from one experiment. The five columns show x and y position, altitude, speed, body deviation angle, and acceleration. The shaded region highlights the dynamic maneuver during which flight metrics (speed, positional error, etc) are computed from. (A-C) and (D-F) correspond to 200 Hz and 100 Hz feedback rates, respectively. }\label{figsup:multi_rate_200_100}
\end{figure}
\newpage

\begin{figure}[H]
\centering
\includegraphics[width=180mm]{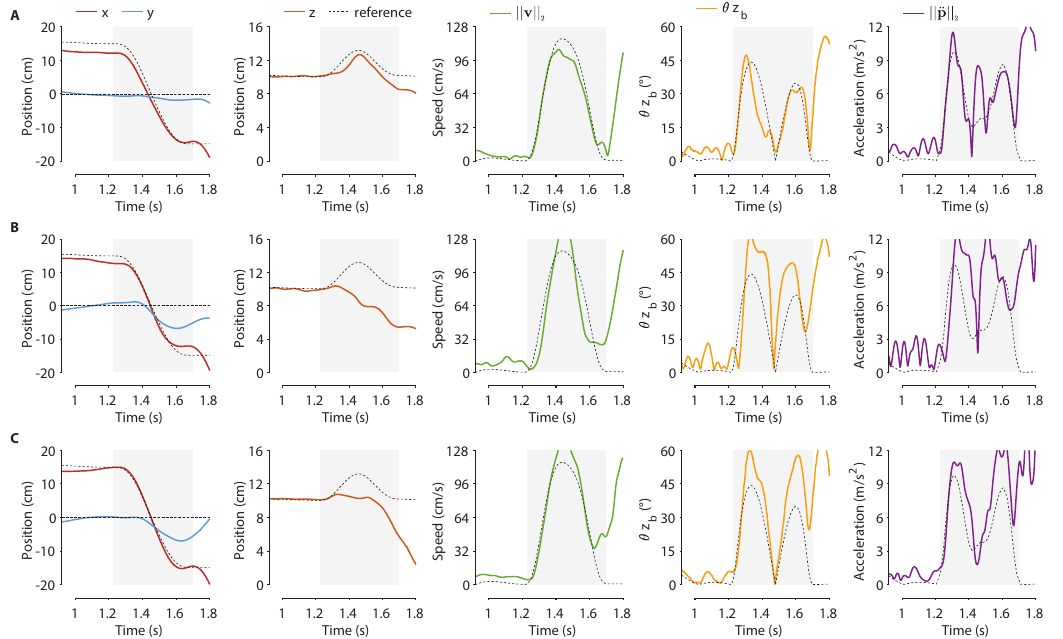}
\vspace*{-0.5cm}
\caption{\textbf{Point-to-point flight demonstrations under 50 Hz feedback rate.} Each horizontal panel (\textbf{A}-\textbf{C}) shows data from one experiment. The five columns show x and y position, altitude, speed, body deviation angle, and acceleration. The shaded region highlights the dynamic maneuver during which flight metrics (speed, positional error, etc) are computed from. }\label{figsup:multi_rate_50}
\end{figure}
\newpage

\begin{figure}[H]
\centering
\includegraphics[width=180mm]{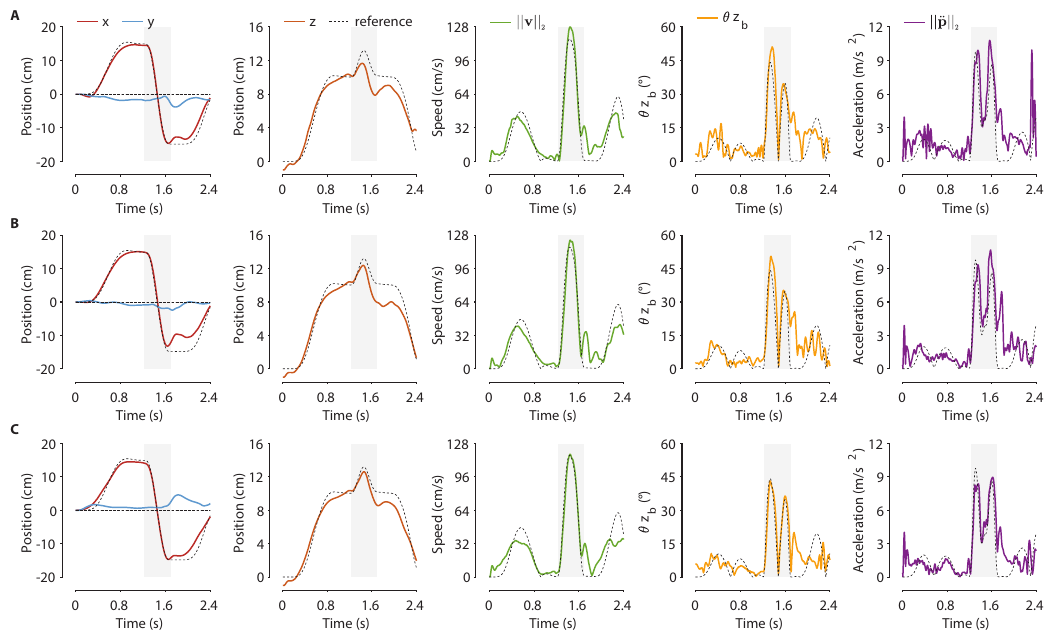}
\vspace*{-0.5cm}
\caption{\textbf{Three point-to-point flight demonstrations with a NN that has 8-neurons.} Each horizontal panel (\textbf{A}-\textbf{C}) shows data from one experiment. The five columns show x and y position, altitude, speed, body deviation angle, and acceleration. The shaded region highlights the dynamic maneuver during which flight metrics (speed, positional error, etc) are computed from.}\label{figsup:multi_size}
\end{figure}
\newpage


\subsection*{Table S1. Robot parameters for constructing the dynamical model }
\begin{table}[ht]
\renewcommand{\arraystretch}{1.5}
\centering
\begin{tabular}{ | m{6.5cm} | m{2cm} | m{3.5cm} | } 
  \hline
  \textbf{Parameter} & \textbf{Symbol} & \textbf{Value} \\
  \hline
  Robot mass & $m$ & 800 mg \\
  \hline
  Robot moment of inertia w.r.t. x-axis & $I_{xx}$ & 43800 mg$\cdot$mm$^2$ \\
  \hline
  Robot moment of inertia w.r.t. y-axis & $I_{yy}$ & 43800 mg$\cdot$mm$^2$ \\
  \hline
  Robot moment of inertia w.r.t. z-axis & $I_{zz}$ & 87600 mg$\cdot$mm$^2$ \\\hline
  Moment arm to x-axis & $l_x$ & 15 mm \\
  \hline
  Moment arm to y-axis & $l_x$ & 15 mm \\
  \hline
  Translational linear drag coefficient & $c_{d,\text{tra}}$ & 0.002 N$\cdot$s/m \\
  \hline
  Rotational quadratic drag coefficient & $c_{d,\text{rot}}$ & 6$\times$10$^{-9}$ N$\cdot$m$\cdot$s$^2$/rad$^2$  
  \\\hline

\end{tabular}
\end{table}

\newpage


\subsection*{Captions for Movies S1 to S7} \label{sec:sup_video}
\vspace{5mm}

\noindent
\textbf{Movie S1: Overview of controller design and robot performance.} The video provides an overview of technical challenges related to insect flight, the design and implementation of a deep-learned robust tube model predictive controller, and the flight demonstrations enabled by this flight controller. 
\vspace{5mm}

\noindent
\textbf{Movie S2: Point-to-point flight demonstration.} The robot flies aggressively between two setpoints that are separated by 30 cm. This maneuver features high translational acceleration of and 11.4 cm/s$^2$. The robot maximum flight speed and body deviation angle are 124 m/s and 49.1°, respectively.
\vspace{5mm}

\noindent
\textbf{Movie S3: Repeated point-to-point flight demonstration.} The robot repeatedly switches between a pair of setpoints four times. Part 1 shows a flight demonstration in which the robot is well calibrated. Part 2 shows the same demonstration under a 33\% command-to-thrust mapping error. Part 3 repeats the same demonstration under a 1.6 m/s wind gust. In part 3, the camera is placed at a 45° angle with respect to the robot’s lateral trajectory. 
\vspace{5mm}

\noindent
\textbf{Movie S4: High acceleration flight demonstration.} The robot tracks 2D trajectories that feature multiple sharp turns. Part 1 shows an X-shaped trajectory where the robot makes nine aggressive turns in 5.5 seconds. Part 2 shows a ``+"-shaped trajectory where the robot compensates for unmodelled downwash during fast descent. 
\vspace{5mm}

\noindent
\textbf{Movie S5: High velocity flight demonstration.} The robot tracks smooth 2D trajectories under high flight speeds. Part 1 shows the top view video of the robot tracking a planar circular path along the xy-plane. The average flight speeds are 153 cm/s, respectively. Part 2 shows the side view video of the robot tracking a figure 8 trajectory along the xz-plane.  The average and maximum flight speeds are 109 and 197 cm/s, respectively. 
\vspace{5mm}

\noindent
\textbf{Movie S6: Body flip demonstration.} The robot performs aerobatic body flip trajectories. Part 1 shows a single body flip demonstration. The maximum linear and angular speeds are 108 cm/s and 1946 °/s, respectively. Part 2 shows the robot completes 10 consecutive flips in a 11-second flight demonstration. The robot rejects disturbances created by its power tether that wraps around the airframe during the second and sixth flips. 
\vspace{5mm}

\noindent
\textbf{Movie S7:  Flight recovery under large disturbance.} A flight experiment in which the robot recovers flight stability after experiencing large disturbances during body flip maneuvers. Due to unmodeled tension from the power tether, the robot states deviate outside the tube that the neural network controller is trained from. The controller detects this event and switches on a model-based controller to stabilize and land the robot.